\documentclass[a4paper]{article}
\usepackage[utf8]{inputenc}
\usepackage{geometry}
\usepackage{microtype}
\usepackage{graphicx}
\usepackage{subfigure}
\usepackage{booktabs} 
\usepackage{microtype}
\usepackage{graphicx,stfloats}
\usepackage{subfigure}
\usepackage{booktabs} 
\usepackage{amsmath,amssymb,amsfonts}
\allowdisplaybreaks
\usepackage{mathtools}
\usepackage{amsthm}
\allowdisplaybreaks[4]
\usepackage{float,bbm}
\usepackage{titling}
\usepackage{setspace}
\usepackage{algorithm,algorithmic}

\makeatletter
\newenvironment{breakablealgorithm}
{
	\begin{center}
		\refstepcounter{algorithm}
		\hrule height.8pt depth0pt \kern2pt
		\renewcommand{\caption}[2][\relax]{
			{\raggedright\textbf{\ALG@name~\thealgorithm} ##2\par}%
			\ifx\relax##1\relax 
			\addcontentsline{loa}{algorithm}{\protect\numberline{\thealgorithm}##2}%
			\else 
			\addcontentsline{loa}{algorithm}{\protect\numberline{\thealgorithm}##1}%
			\fi
			\kern2pt\hrule\kern2pt
		}
	}{
		\kern2pt\hrule\relax
	\end{center}
}
\makeatother
\usepackage{hyperref}
\usepackage[capitalize,noabbrev]{cleveref}

\theoremstyle{plain}
\newtheorem{theorem}{Theorem}[section]
\newtheorem{proposition}{Proposition}[section]
\newtheorem{lemma}{Lemma}[section]
\newtheorem{corollary}{Corollary}[section]
\theoremstyle{definition}

\theoremstyle{remark}
\newtheorem{remark}{Remark}[section]
\usepackage[textsize=tiny]{todonotes}
\hypersetup{hidelinks}

\begin{document}

\begin{center}
	{\Large On the Finite-Time Performance of the Knowledge Gradient Algorithm}
	\\\vspace{20pt}
	Yanwen Li$^1$~~\&~~Siyang Gao$^{1,2}$\\
	{\small \url{yanwen.li@my.cityu.edu.hk}}~~~~{\small \url{siyangao@cityu.edu.hk}}
	\\\hspace{10pt}
	
	\small  
	$^1$ Department of Advanced Design and Systems Engineering, City University of Hong Kong, Hong Kong\\
	$^2$ School of Data Science, City University of Hong Kong, Hong Kong
\end{center}

\vspace{10pt}

\begin{abstract}
	The knowledge gradient (KG) algorithm is a popular and effective algorithm for the best arm identification (BAI) problem. Due to the complex calculation of KG, theoretical analysis of this algorithm is difficult, and existing results are mostly about the asymptotic performance of it, e.g., consistency, asymptotic sample allocation, etc. In this research, we present new theoretical results about the finite-time performance of the KG algorithm. Under independent and normally distributed rewards, we derive bounds for the sample allocation of the algorithm. With these bounds, existing asymptotic results become simple corollaries. Furthermore, we derive upper and lower bounds for the probability of error and simple regret of the algorithm, and show the performance of the algorithm for the multi-armed bandit (MAB) problem. These developments not only extend the existing analysis of the KG algorithm, but can also be used to analyze other improvement-based algorithms. Last, we use numerical experiments to compare the bounds we derive and the performance of the KG algorithm.
\end{abstract}

\vspace{5pt}

\section{Introduction}\label{sec1}

In the best arm identification (BAI) problem, there is a finite number of arms with unknown mean rewards. In each round, an agent chooses an arm to pull and observes a noisy reward. The reward is drawn from a fixed but unknown underlying distribution corresponding to the pulled arm, and no information about other arms is obtained. After learning the mean rewards of the arms by pulling them, the agent identifies an arm that is expected to be the one with the largest mean reward. BAI is also known as the pure exploration problem \cite{Bubeck2009} and serves as a fundamental and useful model for many practical problems such as the inventory management \cite{mahajan2001}, mobile communication \cite{Audibert2010}, A/B testing \cite{scott2015}, and clinic trials \cite{villar2015}.

We consider the BAI problem under the fixed-budget setting, where the agent tries to identify the best arm under a fixed number of rounds. That is, the agent aims to make the maximum use of the available resources (i.e., a limited number of pulls) to explore the set of arms and optimize the quality of the selected arm. The fixed-budget BAI has been widely studied in the literature. Some well-known methods for it include the knowledge gradient (KG, \cite{gupta1996,frazier2008}), expected improvement (EI, \cite{jones1998,ryzhov2016}), optimal computing budget allocation (OCBA, \cite{chen2000a,gao2015tac,gao2017,li2022}), Upper Confidence Bound Exploration (UCB-E, \cite{Audibert2010}), successive rejects (SR, \cite{Audibert2010,bubeck2013}), gap-based exploration (GapE, \cite{gabillon2011}), top-two Thompson sampling (TTTS, \cite{russo2016}), etc.

In this research, we focus on the KG algorithm. It is a single-step Bayesian look-ahead algorithm that was first introduced in \cite{gupta1996}, and was intensively studied later in \cite{frazier2008}. In each round of the KG algorithm, the agent pulls the arm with the largest knowledge gradient, i.e., the arm with the largest expected increment in the posterior mean reward. Such a myopic heuristic is optimal if only one round is left before the agent identifies the best arm. For the general situation where more than one rounds are available, the KG algorithm has also demonstrated excellent empirical performance in various numerical tests, e.g., in \cite{frazier2008}, \cite{frazier2009thesis}, \cite{powell2011}, and \cite{wang2018}. Now, the KG algorithm has been successfully applied to different real problems such as the drug discovery \cite{negoescu2011}, urban delivery \cite{huang2019}, risk quantification \cite{cakmak2020}, and the experimental design in material science \cite{chen2015} and biotechnology \cite{li2018}. The KG algorithm was also extended to solve other types of BAI problems such as the parallel BAI \cite{wu2016} and contextual bandits \cite{ding2021}.

Although the finite-time performance of many BAI algorithms has been well understood \cite{Audibert2010,gabillon2011}, the KG algorithm remains largely undeveloped in this regard. The main reason is that the knowledge gradient function is typically nonlinear and non-convex, which makes it difficult to analyze the algorithm dynamics and thus its performance. Existing literature on the theoretical development of the algorithm is basically limited to its asymptotic performance. \cite{frazier2008} showed that the KG algorithm is consistent, i.e., it is guaranteed to identify the best arm as the number of rounds goes to infinity. \cite{ryzhov2016} derived the sampling rates of each arm (sample allocation) of the KG algorithm. However, in practice, the sample size is finite and might not lead the algorithm to show its large-sample characteristics. The only paper that seeks to analyze the finite-time performance of the KG algorithm, to the best of our knowledge, is \cite{wang2018}. However, their study was conducted based on the submodular assumption on the value of information, which cannot be verified in general, and there exist instances of the BAI problem that violate this assumption. In addition, their target is the worst-case performance of the KG algorithm, and it is presented as a ratio compared to the optimal performance of the algorithm. Nevertheless, this optimal performance is unknown for real problems and can hardly be estimated, so the worst-case performance bound cannot be calculated.

In this paper, we study the finite-time performance of the KG algorithm under very mild conditions. Assuming independent and normally distributed rewards with known variances, we evaluate the performance of the KG algorithm by two common objective measures for BAI: the probability of error (PE, \cite{Audibert2010,bubeck2013icml,kaufmann2016}) and simple regret (SR, \cite{Bubeck2009,gabillon2011,gabillon2012}). PE is the probability that the final recommended arm is not the best one. SR is the difference in the mean reward between the final recommended arm and the best one. We derive upper and lower bounds of PE and SR, corresponding to the worst and best possible performance of the algorithm under the two measures. With the bounds, existing asymptotic results of the algorithm become simple corollaries. Furthermore, our analysis might be extended to other improvement-based methods of BAI, such as the probability improvement (PI, \cite{kashner1964}), expected improvement (EI, \cite{jones1998}), etc.

In addition to BAI, another common bandit model is the multi-armed bandit (MAB). MAB is similar to BAI in that the agent needs to pull an arm in each round and observes a reward of it, but different from BAI, MAB concerns the mean reward obtained in each round \cite{Auer2002,Bubeck2012}. A typical formulation of MAB is minimizing the measure of cumulative regret (CR), which is defined as the sum of the differences in the mean reward between the best arm and each pulled one. For a complete treatment to the KG algorithm, in this research, we also study its finite-time performance under CR.

Last, we conduct numerical experiments to compare the upper and lower bounds with the performance of the KG algorithm.

\subsection*{Organization of the Paper}

Section \ref{sec2} introduces the BAI problem, the KG algorithm, and the three measures we use to evaluate the performance of the algorithm. Theoretical results and some discussion are provided in Section \ref{sec3}; detailed proofs are included in the supplementary material. Section \ref{sec4} performs numerical experiments to illustrate the bounds derived. Section \ref{sec5} concludes this paper and points out future research directions.


\section{Preliminaries}\label{sec2}

In this section, we introduce the BAI problem, the KG algorithm, and the measures for evaluating the performance of the algorithm.

\subsection{Problem Setup}\label{sec2.1}

Let $\left\{1,2,\dots,k\right\}$ be the set of $k$ arms. In each round, the agent chooses an arm to pull. If arm $i$ is pulled in round $t$, we observe a stochastic reward $X_{i,t}$ that follows the normal distribution $\mathcal{N}(\mu_i, \sigma_i^2)$, where the mean $\mu_i$ is unknown and the variance $\sigma_i^2$ is assumed to be known, $i=1,2,\dots,k$. We further assume that the reward observations $X_{i,t}$'s are independent across different arms $i$ and rounds $t$. Suppose there are $n$ rounds in total. Let $I_t$ be the arm pulled by the agent in round $t$, and $J_n$ be the arm recommended by the agent after $n$ rounds. Arm $J_n$ is expected to be the best arm $b\in\arg\max_{i\in\left\{1,\dots,k\right\}}\mu_i$. In this research, we assume that the best arm is unique, i.e., $\mu_b>\mu_i$ for $\forall i\ne b$.

Under a Bayesian framework, the unknown mean $\mu_i$ is treated as a random variable $\tilde{\mu}_{i}$ whose prior distribution is given by $\mathcal{N}\left(\theta_{i,1},\lambda_{i,1}^{2}\right)$, $i=1,\dots,k$. We use the non-informative prior to each $\tilde{\mu}_i$, i.e., $\theta_{i,1}=0$ and $\lambda_{i,1}^2=\infty$ \cite{deGroot1970}. Given the sequence $\left(I_1,X_{I_1,1},\dots,I_t,X_{I_t,t}\right)$, the posterior distribution of $\tilde{\mu}_i$ in round $t$ is $\mathcal{N}\left(\theta_{i,t},\lambda_{i,t}^{2}\right)$ with
\begin{equation}\label{bayes-update1}
	\theta_{i,t+1}=\left\{
	\begin{aligned}
		&\frac{\lambda_{i,t}^{-2}\theta_{i,t}+\sigma_{i}^{-2}X_{i,t}}{\lambda_{i,t}^{-2}+\sigma_{i}^{-2}},~\text{if}~I_t=i,\\
		&\theta_{i,t},~~~~~~~~~~~~~~~~~~~~~~~~~\text{otherwise},
	\end{aligned}
	\right.
\end{equation}
\begin{equation}\label{bayes-update2}
	\lambda_{i,t+1}^{2}=\left\{
	\begin{aligned}
		&\left(\lambda_{i,t}^{-2}+\sigma_{i}^{-2}\right)^{-1},~\text{if}~I_t=i,\\
		&~\lambda_{i,t}^{2},~~~~~~~~~~~~~~~~~~~~\text{otherwise}.
	\end{aligned}
	\right.
\end{equation}
Denote by $\mathcal{S}_t=\left\{\left.\left(\theta_{i,t},\lambda_{i,t}^{2}\right)\right|i=1,\dots,k\right\}$ the knowledge in round $t$. A policy $\varkappa=\left\{I_1,I_2,\dots, I_{n}\right\}$ for BAI corresponds to a sampling rule over the arm set, denoted by $\mathcal{I}^{\varkappa}$. It maps the knowledge $\mathcal{S}_t$ in round $t$ to an arm $\mathcal{I}^{\varkappa}\left(\mathcal{S}_t\right)\in\left\{1,\dots,k\right\}$ that is pulled in round $t$, $t\in\mathbb{N}$. Denote by $\Pi$ the set of policies. In BAI, the goal of the agent is to find the optimal policy, which is the solution to
\begin{align}\label{r&s-dp}
	\sup_{\varkappa\in\Pi}\mathbb{E}^{\varkappa}\left[\max_{i=1,\dots,k}\theta_{i,n}\right],
\end{align} 
where $\mathbb{E}^{\varkappa}$ denotes a conditional expectation given $I_{t}=\mathcal{I}^{\varkappa}\left(\mathcal{S}_{t}\right)$ for $t=1,\dots,n$.

\subsection{Knowledge Gradient}\label{sec2.2}

The KG algorithm is simple in concept. Iteratively, it assumes that we only have one sample left and pull the arm that maximizes the expectation of the single-period increase in $\max_i\theta_{i,t}$. In other words, the algorithm tries to maximize the expectation of $\max_i\theta_{i,t+1}$ after round $t$, where $\theta_{i,t+1}$ for each arm $i$ is treated as a random variable before round $t+1$ since the reward in round $t+1$ is unknown in round $t$, $t=1,\dots,n-1$. To this end, we can provide the following formula $v_{i,t}^{\text{KG}}$ to compute the expected increment of $\theta_{i,t}$ after round $t$,
\begin{align}\label{kg_value_func}
	v_{i,t}^{\text{KG}}=\mathbb{E}\left[\left.\max_{i'}\theta_{i',t+1}-\max_{i'}\theta_{i',t}\right|I_t=i,\mathcal{S}_t\right],
\end{align}
and write the sampling rule of the KG algorithm $\mathcal{I}^{\rm KG}$ as
\begin{align}\label{kg_decision_rule}
	\mathcal{I}^{\rm KG}\left(\mathcal{S}_t\right)=\mathop{\arg\max}_{i=1,\dots,k}v_{i,t}^{\text{KG}}.
\end{align}
The acquisition function (\ref{kg_value_func}) does not have an analytical expression. To handle the difficulty, \cite{gupta1996} and \cite{frazier2008} provided reasonable approximations for developing computationally tractable algorithms. In the approximations, $v_{i,t}^{\text{KG}}$ in (\ref{kg_value_func}) is replaced by the following, 
\begin{equation}\label{value-kg}
	\begin{aligned}
		v_{i,t}^{\text{KG}}=&\zeta_{i,t}f\left(-\frac{\left|\theta_{i,t}-\max_{j\ne i}\theta_{j,t}\right|}{\zeta_{i,t}}\right),
	\end{aligned}
\end{equation}
where $\zeta_{i,t}^{2}={\rm Var}\left(\left.\theta_{i,t+1}\right|I_t=i,\mathcal{S}_t\right)=\lambda_{i,t}^{2}-\lambda_{i,t+1}^{2}$ can be interpreted as the variance of the change in $\theta_{i,t+1}-\theta_{i,t}$ resulting from the next pull since ${\rm Var}\left[\left.\theta_{i,t+1}\right|I_t=i,\mathcal{S}_t\right]={\rm Var}\left[\left.\theta_{i,t+1}-\theta_{i,t}\right|I_t=i,\mathcal{S}_t\right]$. Function $f(x)=x\Phi(x)+\phi(x)$ is a monotone increasing function with respect to $x$, and $\Phi(x)$ and $\phi(x)$ are the cumulative density function and the probability density function of the standard normal distribution. We summarize the KG algorithm as follows.

\begin{breakablealgorithm}
	\caption{Knowledge Gradient}\label{alg-kg}
	\begin{algorithmic}
		\STATE {\bfseries Input:} number of arms $k$, number of rounds $n$.
		\STATE Initialize $\mathcal{S}_1=\left\{\left.\left(\theta_{i,1},\lambda_{i,1}^{2}\right)\right|i=1,\dots,k\right\}$ and set $N_{1,1}=\dots=N_{k,1}=0$.
		\FOR{$t=1$ {\bfseries to} $n-1$}
		\STATE Compute $I_{t}=\mathop{\arg\max}_{i} v_{i,t}^{\rm KG}$ based on (\ref{value-kg}).
		\STATE Observe a reward $X_{I_{t},t}\sim\mathcal{N}\left(\mu_{I_{t}},\sigma_{I_{t}}^{2}\right)$.
		\STATE Compute $\theta_{i,t+1}$ and $\lambda_{i,t+1}^{2}$ based on (\ref{bayes-update1}) and (\ref{bayes-update2}), and update the knowledge set $\mathcal{S}_{t+1}$.
		\STATE $N_{I_{t},t+1}=N_{I_{t},t}+1$, $N_{i,t+1}=N_{i,t}$ for $\forall i\ne I_{t}$. 
		\STATE $t\leftarrow t+1$.
		\ENDFOR
		\STATE {\bfseries Output:} $J_n=\arg\max_i\theta_{i,n}$.
	\end{algorithmic}
\end{breakablealgorithm}

The asymptotic performance of the KG algorithm has been studied in the literature. \cite{frazier2008} showed that with the algorithm, $J_n$ converges to the best arm $b$ as $n\to\infty$, i.e., the consistency. \cite{ryzhov2016} showed that the sampling rate $\left\{\left.\alpha_{i,t}=\frac{N_{i,t}}{t}\right|t\in\mathbb{N}\right\}$ of each arm $i\in\left\{1,\dots,k\right\}$ generated by the algorithm satisfies
\begin{equation}\label{sampleallocation-kg}
	\begin{aligned}
		&\lim_{t\to\infty}\frac{\alpha_{i_1,t}}{\alpha_{i_2,t}}\overset{a.s.}{=}\frac{\sigma_{i_1}\left(\mu_b-\mu_{i_2}\right)}{\sigma_{i_2}\left(\mu_b-\mu_{i_1}\right)},~\forall i_1,i_2\ne b,\\
		&\lim_{t\to\infty}\frac{\alpha_{i,t}}{\alpha_{b,t}}\overset{a.s.}{=}\frac{\sigma_i\left(\mu_b-\max_{i\ne b}\mu_i\right)}{\sigma_b\left(\mu_b-\mu_i\right)},~\forall i\ne b,
	\end{aligned}
\end{equation}
where $N_{i,t}=\sum_{s=1}^{t}\mathbbm{1}\left\{I_s=i\right\}$ is the number of pulls allocated to arm $i$ after round $t$, the indicator function $\mathbbm{1}\left\{\cdot\right\}$ equals one if its argument is true and is zero otherwise, and ``a.s.'' means ``almost surely''.

\subsection{Performance Measures}\label{sec2.3}

In contrast to \cite{wang2018} which evaluates the performance of the KG algorithm by a self-defined measure, we conduct our analysis under three common measures in BAI and MAB. Below we show the three measures under a frequentist framework where $\mu_i$ of each arm $i$ has a fixed but unknown value.
\begin{itemize}
	\item Probability of error (PE). PE is an important performance measure for BAI \cite{Audibert2010,bubeck2013icml,kaufmann2016}. It is the probability that the estimated best arm $J_n$ is not the true best one. PE can also be treated as the expectation of the 0-1 loss function $\mathbbm{1}\left\{J_n\ne b\right\}$, and the expectation is taken over the sequence $\left(I_1,X_{I_1,1},\dots,I_t,X_{I_t,t}\right)$ whose realizations determine the estimated best arm $J_n$, conditional on the unknown means $\boldsymbol{\mu}=\left(\mu_1,\dots,\mu_k\right)$. In this research, the PE is denoted by
	\begin{align*}
		e_n=\mathbb{P}\left(\left.J_n\ne b\right|\boldsymbol{\mu}\right).
	\end{align*}
	
	\item Simple regret (SR). SR is another important measure for BAI \cite{Bubeck2009,Audibert2010,gabillon2011,gabillon2012}. It is the expectation of the differences in the mean reward between the true best arm and the estimated best one. In this research, the SR is denoted by 
	\begin{align*}
		r_n=\mathbb{E}\left[\left.\mu_b-\mu_{J_n}\right|\boldsymbol{\mu}\right].
	\end{align*}
	Sometimes SR is also called opportunity cost \cite{scott2015,gao2017} or linear loss \cite{chick2010}. 
	
	\item Cumulative regret (CR). CR is mostly used for the MAB problem \cite{Auer2002,Bubeck2012}. It is defined as the sum of the differences in the mean reward between the true best arm and each pulled one. The CR after $n$ rounds is given by 
	\begin{align*}
		R_n=\sum_{i\ne b}\left(\mu_b-\mu_i\right)\mathbb{E}\left[\left.N_{i,t}\right|\boldsymbol{\mu}\right].
	\end{align*}
\end{itemize}

\section{Theoretical Results}\label{sec3}

Although the KG algorithm was initially proposed in a Bayesian setting where we have an initial belief to the mean $\mu_i$ of each arm $i$, when we adopt the non-informative prior, parameters $\theta_{i,t+1}$ and $\lambda_{i,t+1}^{2}$ in (\ref{bayes-update1}) and (\ref{bayes-update2}) are updated in the same manner as in the frequentist setting. In this section, we will conduct our analysis under the frequentist setting.

Let $\hat{\mu}_{i,t}=\frac{1}{N_{i,t}}\sum_{s=1}^{t}\mathbbm{1}\left\{I_s=i\right\}X_{i,s}$ denote the estimated mean reward of arm $i$ after round $t$. Notice that for $\forall t$, $\forall i$, with given rewards $X_{i,1},\dots,X_{i,t}$, $\hat{\mu}_{i,t}=\theta_{i,t}$ for $\theta_{i,t}$ in (\ref{bayes-update1}). Under the frequentist setting, we treat $\hat{\mu}_{i,t}$ as a random variable. For $\forall t$, $\forall i$, given that $N_{i,t}=n_i$, $\hat{\mu}_{i,t}$ follows a normal distribution with mean $\mu_i$ and variance $\frac{\sigma_i^2}{n_i}$ \cite{dekking2005}.

Below, we first provide some lemmas to facilitate our analysis. Lemma \ref{lemma1} gives a bound on the difference between the estimated mean $\hat{\mu}_{i,t}$ and the true mean $\mu_i$ of each arm $i\in\left\{1,\dots,k\right\}$ under independent and normally distributed rewards. Lemma \ref{lemma2} shows a concentration inequality in the case of normal distribution. Lemma \ref{lemma3} provides an upper bound and a lower bound of the function $f(x)=x\Phi(x)+\phi(x)$. Lemma \ref{lemma4} is a proposition of Bernstein's maximal inequality for martingales.

\begin{lemma} \label{lemma1}
	(Lemma 5 of \cite{qin2017}) Under any sampling rule and the non-informative prior for each arm, there exists a random variable $W$ that depends on the sampling rule and satisfies that $\mathbb{E}\left[e^{\gamma W}\right]<\infty$ for $\forall \gamma>0$, and it holds a.s. that for $\forall i\in\left\{1,\dots,k\right\}$,
	\begin{align*}
		\left|\hat{\mu}_{i,t}-\mu_i\right|\leq\sigma_{i}W\sqrt{\frac{\log\left(e+N_{i,t}\right)}{1+N_{i,t}}},~\forall t\in\mathbb{N}.
	\end{align*}
\end{lemma}

\begin{lemma}\label{lemma2}
	Suppose stochastic rewards $Y_1,\dots,Y_m\sim\mathcal{N}\left(\mu,\sigma^{2}\right)$. Denote by $\hat{Y}_m=\frac{1}{m}\sum_{s=1}^{m}Y_s$. For $\forall\epsilon\geq 0$, 
	\begin{align*}
		\mathbb{P}\left(\left|\hat{Y}_m-\mu\right|\geq \epsilon\right)\leq\frac{2\sigma}{\sqrt{m}\epsilon}\exp\left\{-\frac{m\epsilon^{2}}{2\sigma^{2}}\right\}.
	\end{align*}
\end{lemma}

\begin{lemma}\label{lemma3}
	For function $f(x)=x\Phi(x)+\phi(x)$, if $x\geq 2$, then 
	\begin{align*}
		\frac{\phi(x)}{x^{3}}<f(-x)<\frac{\phi(x)}{x^{2}},
	\end{align*}
	where $\Phi(x)=\int_{-\infty}^{x}\phi(r){\rm d}r$, $\phi(x)=\frac{1}{\sqrt{2\pi}}e^{-\frac{x^{2}}{2}}$.
\end{lemma}

\begin{lemma}\label{lemma4}
	(Lemma 1 of \cite{cesa2008}) Denote by $\left\{L_s\big{|}0\leq L_s\leq 1,s=1,2,\dots\right\}$ a sequence of random variables.
	\begin{itemize}
		\item Define the bounded martingale difference sequence $M_s=\mathbb{E}\left[L_s\big{|}L_1,\dots,L_{s-1}\right]-L_s$ and the associated martingale $K_t=M_1+\dots+M_t$ with conditional variance $V_t=\sum_{s=1}^{t}{\rm Var}\left[L_s\big{|}L_1,\dots,L_{s-1}\right]$. For any $\kappa,\omega\geq 0$, 
		\begin{align*}
			\mathbb{P}\left(K_t\geq\kappa,V_t\leq\omega\right)\leq \exp\left\{-\frac{\kappa^{2}}{2\omega+\frac{2\kappa}{3}}\right\}.
		\end{align*}
		
		\item Define another bounded martingale difference sequence $\tilde{M}_s=-M_s$ and the associated martingale $\tilde{K}_t=\tilde{M}_1+\dots+\tilde{M}_t$ with conditional variance $V_t=\sum_{s=1}^{t}{\rm Var}\left[L_s\big{|}L_1,\dots,L_{s-1}\right]$. Then, for any $ \tilde{\kappa},\tilde{\omega}\geq 0$, 
		\begin{align*}
			\mathbb{P}\left(\tilde{K}_t\geq\tilde{\kappa},V_t\leq\tilde{\omega}\right)\leq \exp\left\{-\frac{\tilde{\kappa}^{2}}{2\tilde{\omega}+\frac{2\tilde{\kappa}}{3}}\right\}.
		\end{align*}
	\end{itemize}
\end{lemma}

\subsection{Analysis of the Sample Allocation}\label{sec3.1}

We first focus on the sample allocation of the KG algorithm. It serves as a basis for more in-depth analysis of the three performance measures PE, SR, and CR.

Based on Lemma \ref{lemma1}, we show that each arm can be pulled frequently under the KG algorithm.

\begin{proposition}\label{prop1}
	Under the KG algorithm, $\exists T_0$, $\forall t\geq T_0$, it holds a.s. for the number of pulls $N_{i,t}$ of arm $i$ that
	\begin{align*}
		N_{i,t}\geq\left(\frac{t}{k}\right)^{\frac{3}{4}},~\forall i\in\left\{1,\dots,k\right\}.
	\end{align*}
\end{proposition}

Proposition \ref{prop1} establishes a lower bound $\left(\frac{t}{k}\right)^{\frac{3}{4}}$ for the number of pulls of each arm. Since this lower bound goes to infinity as the number of rounds $t$ goes to infinity, the consistency of the KG algorithm immediately follows, i.e., the mean estimates will converge to the true means as $t\to\infty$, and the estimated best arm will converge to the true best arm. To prove it, we can show that if any arm $i_0$ receives too few number of pulls, the value of $v_{i_0,t}^{\rm KG}$ in (\ref{value-kg}) is higher than that of the arms which receive a sufficiently large number of pulls. Then, arm $i_0$ will be pulled in the next round. 

Based on Lemmas \ref{lemma2}, \ref{lemma3}, and Proposition \ref{prop1}, we show an upper bound and a lower bound of $\frac{N_{i,t}}{N_{b,t}}$, $\forall i\ne b$. The bounds hold with a probability converging to one as $t\to\infty$. 

\begin{proposition}\label{prop2}
	Under the KG algorithm, $\exists T>T_0$, it holds with a probability of at least $\left[1-q\left(\frac{3}{4}t\right)\right]^{k}$ that for $\forall i\ne b$, $\forall t>T$,
	\begin{align*}
		\underline{\rho}_{i,b,t}\leq\frac{N_{i,t}}{N_{b,t}}\leq\overline{\rho}_{i,b,t},
	\end{align*}
	where $q(s)=4\sigma_{\rm max} k^{-\frac{1}{8}}s^{-\frac{1}{8}}\exp\left\{-\frac{k^{\frac{1}{4}}s^{\frac{1}{4}}}{8\sigma_{\max}^{2}}\right\}$, $\sigma_{\rm max}=\max_{i\in\left\{1,\dots,k\right\}}\sigma_i$, $\sigma_{\rm min}=\min_{i\in\left\{1,\dots,k\right\}}\sigma_i$, $\delta_{\rm max}=\max_{i_1\ne i_2}\left|\mu_{i_1}-\mu_{i_2}\right|$, $\underline{\rho}_{i,b,t}=\min\left\{\underline{\rho}_{i,b,t}^{(1)},\underline{\rho}_{i,b,t}^{(2)}\right\}$, $\overline{\rho}_{i,b,t}=\max\left\{\overline{\rho}_{i,b,t}^{(1)},\overline{\rho}_{i,b,t}^{(2)}\right\}$,
	\begin{align*}
		\underline{\rho}_{i,b,t}^{(1)}=&\frac{\mu_b-\underset{j\ne b}{\max}~\mu_j-\left(\frac{3}{4}t\right)^{-\frac{1}{4}}}{\left(1+\left(\frac{3t}{4k}\right)^{-\frac{3}{4}}\right)^{2}\sigma_b}\left(\frac{\left(\mu_b-\mu_i+\left(\frac{3}{4}t\right)^{-\frac{1}{4}}\right)^{2}}{\sigma_i^{2}}+\frac{16k}{\sqrt{3t}}+\frac{8k\max\left\{\ln\left(\frac{27\delta_{\rm max}^{3}}{8\sigma_{\rm min}^{4}}\right),0\right\}}{3t}\right)^{-\frac{1}{2}},\\
		\underline{\rho}_{i,b,t}^{(2)}=&\frac{\mu_b-\max_{j\ne b}\mu_j-t^{-\frac{1}{4}}}{\left(1+\left(\frac{t}{k}\right)^{-\frac{3}{4}}\right)\sigma_b}\left(\frac{\left(\mu_b-\mu_i+t^{-\frac{1}{4}}\right)^{2}}{\sigma_i^{2}}+\frac{8k}{\sqrt{t}}+\frac{2k\max\left\{\ln\left(\frac{27\delta_{\rm max}^{3}}{8\sigma_{\rm min}^{4}}\right),0\right\}}{t}\right)^{-\frac{1}{2}},\\
		\overline{\rho}_{i,b,t}^{(1)}=&\frac{\left(1+\left(\frac{t}{k}\right)^{-\frac{3}{4}}\right)\sigma_i}{\mu_b-\mu_i-t^{-\frac{1}{4}}}\left(\frac{\left(\mu_b-\max_{j\ne b}\mu_j+t^{-\frac{1}{4}}\right)^{2}}{\sigma_b^{2}}+\frac{8k}{\sqrt{t}}+\frac{2k\max\left\{\ln\left(\frac{27\delta_{\rm max}^{3}}{8\sigma_{\rm min}^{4}}\right),0\right\}}{t}\right)^{\frac{1}{2}},\\
		\overline{\rho}_{i,b,t}^{(2)}=&\frac{\left(1+\left(\frac{3t}{4k}\right)^{-\frac{3}{4}}\right)^{2}\sigma_i}{\mu_b-\mu_i-\left(\frac{3}{4}t\right)^{-\frac{1}{4}}}\left(\frac{\left(\mu_b-\underset{j\ne b}{\max}~\mu_j+\left(\frac{3}{4}t\right)^{-\frac{1}{4}}\right)^{2}}{\sigma_b^{2}}+\frac{16k}{\sqrt{3t}}+\frac{8k\max\left\{\ln\left(\frac{27\delta_{\rm max}^{3}}{8\sigma_{\rm min}^{4}}\right),0\right\}}{3t}\right)^{\frac{1}{2}}.
	\end{align*}
\end{proposition}

With Proposition \ref{prop2}, we can derive an upper bound and a lower bound of the sampling rate $\alpha_{i,t}$, $\forall i\in\left\{1,\dots,k\right\}$, i.e., the proportion of pulls allocated to arm $i$ until round $t$. The sampling ratio $N_{i,t}/N_{b,t}$ in Proposition \ref{prop2} is closely related to the sampling rate $\alpha_{i,t}$ in Theorem 1 because $\alpha_{i,t}=\frac{N_{i,t}/N_{b,t}}{\sum_{j=1}^{k}N_{j,t}/N_{b,t}}$, $\forall i$. Proposition \ref{prop2} provides analytical upper and lower bounds of the sampling ratio $N_{i,t}/N_{b,t}$. With these bounds, we can replace $N_{i,t}/N_{b,t}$ in the numerator of $\alpha_{i,t}=\frac{N_{i,t}/N_{b,t}}{\sum_{j=1}^{k}N_{j,t}/N_{b,t}}$ by its upper bound and replace $N_{j,t}/N_{b,t}$ in the denominator of $\alpha_{i,t}=\frac{N_{i,t}/N_{b,t}}{\sum_{j=1}^{k}N_{j,t}/N_{b,t}}$ by its lower bound. In this way, we can obtain the upper and lower bounds of the sampling rates $\alpha_{i,t}$ in Theorem \ref{thm1}, $i=1,\dots,k$. The bounds hold with a probability converging to one as $t\to\infty$.

\begin{theorem}\label{thm1}
	Under the KG algorithm, $\exists T>T_0$, it holds with a probability of at least $\left[1-q\left(\frac{3}{4}t\right)\right]^{k}$ that for $\forall t>T$,
	\begin{align*}
		\frac{1}{1+\sum_{i\ne b}\overline{\rho}_{i,b,t}}\leq& \alpha_{b,t}\leq\frac{1}{1+\sum_{i\ne b}\underline{\rho}_{i,b,t}},\\
		\frac{\underline{\rho}_{i,b,t}}{1+\sum_{j\ne b}\overline{\rho}_{j,b,t}}\leq& \alpha_{i,t}\leq\frac{\overline{\rho}_{i,b,t}}{1+\sum_{j\ne b}\underline{\rho}_{j,b,t}},~\forall i\ne b,
	\end{align*}
	where $q(\cdot)$, $\underline{\rho}_{i,b,t}$, and $\overline{\rho}_{i,b,t}$ are from Proposition \ref{prop2}.
\end{theorem}

Note that the lower and upper bounds of $\alpha_{i,t}$ in Theorem \ref{thm1} converge to the same value which falls in $(0,1)$ as $t$ goes to infinity for $i=1,\dots,k$. It is a more elaborate result than Proposition \ref{prop1} describing the number of pulls of each arm. It implies that the number of pulls of each arm will approximately show a linear increase during the sampling process of the KG algorithm. 

\begin{corollary}\label{cor1}
	Under the KG algorithm,
	\begin{align*}
		&\lim_{t\to\infty}\frac{N_{i,t}}{N_{b,t}}\overset{a.s.}{=}\frac{\sigma_i}{\sigma_b}\frac{\mu_b-\max_{j\ne b}\mu_j}{\mu_b-\mu_i},\forall i\ne b,\\
		&\lim_{t\to\infty}\frac{N_{i_1,t}}{N_{i_2,t}}\overset{a.s.}{=}\frac{\sigma_{i_1}}{\sigma_{i_2}}\frac{\mu_b-\mu_{i_2}}{\mu_b-\mu_{i_1}},\forall i_1,i_2\ne b.
	\end{align*}
\end{corollary}

Corollary \ref{cor1} depicts the asymptotic sample allocation of the KG algorithm. It is a simple corollary of Theorem \ref{thm1}, and can be obtained by analyzing the lower and upper bounds in Theorem \ref{thm1} as $t\to\infty$. Note that this corollary aligns with (\ref{sampleallocation-kg}) that was first shown in \cite{ryzhov2016}.

\subsection{Analysis of PE, SR and CR}\label{sec3.2}

We denote by $J_t$, $e_t$, and $r_t$ the estimated best arm, PE, and SR after round $t$, $t=1,\dots,n$. With Theorem \ref{thm1}, we can characterize the worst-case performance of the PE and SR for the KG algorithm. 

\begin{theorem}\label{thm2}
	Under the KG algorithm, $\exists T>T_0$, $\forall t>T$,
	\begin{itemize}
		\item for PE, 
		\begin{align*}
			e_t\leq&\frac{\sigma_b\sqrt{2\left(1+\sum_{i\ne b}\overline{\rho}_{i,b,t}\right)}}{\delta_{\rm min}\sqrt{\pi t}}\exp\left\{-\frac{\delta_{\rm min}^2}{8\sigma_b^2\left(1+\sum_{i\ne b}\overline{\rho}_{i,b,t}\right)}t\right\}\\
			&+\frac{\sqrt{2} k^{\frac{3}{8}}\sigma_b}{\sqrt{\pi}\delta_{\rm min}t^{\frac{3}{8}}}\left[1-\left[1-q\left(\frac{3}{4}t\right)\right]^{k}\right]\exp\left\{-\frac{\delta_{\rm min}^2}{8\sigma_b^2 k^{\frac{3}{4}}}t^{\frac{3}{4}}\right\}\\
			&+\sum_{i\ne b}\left\{\rule{0em}{10mm}\frac{\sigma_i\sqrt{\left(1+\sum_{i\ne b}\overline{\rho}_{i,b,t}\right)}}{\left(\mu_b-\mu_i-\frac{\delta_{\rm min}}{2}\right)\sqrt{2\pi\underline{\rho}_{i,b,t} t}}\right.\exp\left\{-\frac{\left(\mu_b-\mu_i-\frac{\delta_{\rm min}}{2}\right)^2\underline{\rho}_{i,b,t}}{2\sigma_i^2\left(1+\sum_{i\ne b}\overline{\rho}_{i,b,t}\right)}t\right\}\\
			&~~~~~~~~~~+\frac{k^{\frac{3}{8}}\sigma_i\left[1-\left[1-q\left(\frac{3}{4}t\right)\right]^{k}\right]}{\sqrt{2\pi}\left(\mu_b-\mu_i-\frac{\delta_{\rm min}}{2}\right)t^{\frac{3}{8}}}\left.\exp\left\{-\frac{\left(\mu_b-\mu_i-\frac{\delta_{\rm min}}{2}\right)^2}{2\sigma_i^2k^{\frac{3}{4}}}t^{\frac{3}{4}}\right\}\rule{0em}{10mm}\right\},
		\end{align*}
		\item for SR, 
		\begin{align*}
			r_t\leq&\frac{\delta_{\rm max}\sigma_b\sqrt{2\left(1+\sum_{i\ne b}\overline{\rho}_{i,b,t}\right)}}{\delta_{\rm min}\sqrt{\pi t}}\exp\left\{-\frac{\delta_{\rm min}^2}{8\sigma_b^2\left(1+\sum_{i\ne b}\overline{\rho}_{i,b,t}\right)}t\right\}\\
			&+\frac{\sqrt{2} k^{\frac{3}{8}}\delta_{\rm max}\sigma_b}{\sqrt{\pi}\delta_{\rm min}t^{\frac{3}{8}}}\left[1-\left[1-q\left(\frac{3}{4}t\right)\right]^{k}\right]\exp\left\{-\frac{\delta_{\rm min}^2}{8\sigma_b^2 k^{\frac{3}{4}}}t^{\frac{3}{4}}\right\}\\
			&+\sum_{i\ne b}\left\{\rule{0em}{10mm}\frac{\delta_{\rm max}\sigma_i\sqrt{1+\sum_{i\ne b}\overline{\rho}_{i,b,t}}}{\left(\mu_b-\mu_i-\frac{\delta_{\rm min}}{2}\right)\sqrt{2\pi\underline{\rho}_{i,b,t} t}}\right.\exp\left\{-\frac{\left(\mu_b-\mu_i-\frac{\delta_{\rm min}}{2}\right)^2\underline{\rho}_{i,b,t}}{2\sigma_i^2\left(1+\sum_{i\ne b}\overline{\rho}_{i,b,t}\right)}t\right\}\\
			&~~~~~~~~~~+\frac{k^{\frac{3}{8}}\delta_{\rm max}\sigma_i\left[1-\left[1-q\left(\frac{3}{4}t\right)\right]^{k}\right]}{\sqrt{2\pi}\left(\mu_b-\mu_i-\frac{\delta_{\rm min}}{2}\right)t^{\frac{3}{8}}}\left.\exp\left\{-\frac{\left(\mu_b-\mu_i-\frac{\delta_{\rm min}}{2}\right)^2}{2\sigma_i^2k^{\frac{3}{4}}}t^{\frac{3}{4}}\right\}\rule{0em}{10mm}\right\},
		\end{align*}
	\end{itemize}
	where $q(\cdot)$, $\underline{\rho}_{i,b,t}$, and $\overline{\rho}_{i,b,t}$ are from Proposition \ref{prop2}, $\delta_{\rm min}=\min\left\{\left|\mu_{i_1}-\mu_{i_2}\right|>0\left|i_1\ne i_2\right.\right\}$.
\end{theorem}

We can also derive the lower bounds of PE and SR. They show the best performance the KG algorithm can possibly achieve. 

\begin{proposition}\label{prop3}
	Under the KG algorithm, $\exists T>T_0$, $\forall t>T$, 
	\begin{itemize}
		\item for PE,
		\begin{align*}
		    e_t\geq\frac{\left[1-q\left(\frac{3}{4}t\right)\right]^{2k}}{2\pi}\min_{j\ne b}&\left\{\rule{0em}{10mm}\frac{\frac{\delta_{\rm min}}{2\sigma_j}\sqrt{\frac{\underline{\rho}_{j,b,t}}{1+\sum_{i\ne b}\overline{\rho}_{i,b,t}}}}{1+\frac{\delta_{\rm min}^2\overline{\rho}_{j,b,t}t}{4\sigma_j^2\left(1+\sum_{i\ne b}\underline{\rho}_{i,b,t}\right)}}\frac{\frac{\left(\mu_b-\mu_j-\frac{\delta_{\rm min}}{2}\right)t}{\sigma_b\sqrt{1+\sum_{i\ne b}\overline{\rho}_{i,b,t}}}}{1+\frac{\left(\mu_b-\mu_j-\frac{\delta_{\rm min}}{2}\right)^2 t}{\sigma_b^2\left(1+\sum_{i\ne b}\underline{\rho}_{i,b,t}\right)}}\right.\\
		    &~~\left.\cdot\exp\left\{-\frac{\delta_{\rm min}^2\overline{\rho}_{j,b,t}t}{8\sigma_j^2\left(1+\sum_{i\ne b}\underline{\rho}_{i,b,t}\right)}\right\}\exp\left\{-\frac{\left(\mu_b-\mu_j-\frac{\delta_{\rm min}}{2}\right)^2 t}{2\sigma_b^2\left(1+\sum_{i\ne b}\underline{\rho}_{i,b,t}\right)}\right\}\rule{0em}{10mm}\right\},
		\end{align*}
		\item for SR, 
		\begin{align*}
		r_t\geq\frac{\left[1-q\left(\frac{3}{4}t\right)\right]^{2k}\delta_{\rm min}}{2\pi}&\min_{j\ne b}\left\{\rule{0em}{10mm}\frac{\frac{\delta_{\rm min}}{2\sigma_j}\sqrt{\frac{\underline{\rho}_{j,b,t}}{1+\sum_{i\ne b}\overline{\rho}_{i,b,t}}}}{1+\frac{\delta_{\rm min}^2\overline{\rho}_{j,b,t}t}{4\sigma_j^2\left(1+\sum_{i\ne b}\underline{\rho}_{i,b,t}\right)}}\frac{\frac{\left(\mu_b-\mu_j-\frac{\delta_{\rm min}}{2}\right)t}{\sigma_b\sqrt{1+\sum_{i\ne b}\overline{\rho}_{i,b,t}}}}{1+\frac{\left(\mu_b-\mu_j-\frac{\delta_{\rm min}}{2}\right)^2 t}{\sigma_b^2\left(1+\sum_{i\ne b}\underline{\rho}_{i,b,t}\right)}}\right.\\
		&~~\left.\cdot\exp\left\{-\frac{\delta_{\rm min}^2\overline{\rho}_{j,b,t}t}{8\sigma_j^2\left(1+\sum_{i\ne b}\underline{\rho}_{i,b,t}\right)}\right\}\exp\left\{-\frac{\left(\mu_b-\mu_j-\frac{\delta_{\rm min}}{2}\right)^2 t}{2\sigma_b^2\left(1+\sum_{i\ne b}\underline{\rho}_{i,b,t}\right)}\right\}\rule{0em}{10mm}\right\},
		\end{align*}
	\end{itemize}
	where $q(\cdot)$, $\underline{\rho}_{i,b,t}$, and $\overline{\rho}_{i,b,t}$ are from Proposition \ref{prop2}.
\end{proposition}


%

In addition to the PE and SR, we characterize the performance of the KG algorithm for the MAB problem, under the measure of CR. We denote by $R_t$ the CR after round $t$, $t=1,\dots,n$.

\begin{theorem}\label{thm3}
	Under the KG algorithm, the following statements hold:
	\begin{itemize}
		\item $\exists T>T_0$, after round $t>T$, 
		\begin{align*}
			\hspace{-6mm}
			R_t<\frac{\sum_{i\ne b}\left(\mu_b-\mu_i\right)\overline{\rho}_{i,b,t}}{1+\sum_{i\ne b}\underline{\rho}_{i,b,t}}t+k\sum_{i\ne b}\left(\mu_b-\mu_i\right)q\left(\frac{3}{4}t\right)t.
		\end{align*}
		
		\item $\lim_{t\to\infty}\frac{R_t}{t}=\frac{\sum_{i\ne b}\sigma_i}{\frac{\sigma_b}{\mu_b-\max_{j\ne b}\mu_j}+\sum_{i\ne b}\frac{\sigma_i}{\mu_b-\mu_i}}$.
	\end{itemize}
	where $q(\cdot)$, $\underline{\rho}_{i,b,t}$, and $\overline{\rho}_{i,b,t}$ are from Proposition \ref{prop2}.
\end{theorem}

Theorem \ref{thm2}, Proposition \ref{prop3}, and Theorem \ref{thm3} show that the PE and SR of the KG algorithm converge exponentially fast to zero while CR increases linearly with the number of rounds $t$. This result aligns with the theoretical findings in \cite{Bubeck2009} that an algorithm with a linear growth under CR will have PE and SR converging to zero exponentially fast at best.

\begin{remark}
	Theorem \ref{thm3} shows that CR of the KG algorithm approximately demonstrates a linear increase with the round index $t$. It is suboptimal compared to the optimal logarithmic increasing rate of CR \cite{burnetas1997,Auer2002}. This is reasonable, because the KG algorithm seeks to identify the best arm after $n$ rounds, and was not designed to minimize the cumulative cost incurred by sampling in each round. If we want to modify the KG algorithm for CR to achieve the logarithmic increasing rate, a possible way is to change the acquisition function $v_{i,t}^{\text{KG}}$ from $\zeta_{i,t}f\left(-\frac{\left|\theta_{i,t}-\max_{j\ne i}\theta_{j,t}\right|}{\zeta_{i,t}}\right)$ to $\zeta_{i,t}f\left(-\frac{\left|\theta_{i,t}-\max_{j}\theta_{j,t}\right|}{\zeta_{i,t}}\right)$. Subsequent theoretical development can be made by following similar discussion as used in the proof of Proposition 5 of \cite{ryzhov2016}. Detailed analysis of it is out of scope for this paper.
	\hfill$\blacksquare$
\end{remark}


\begin{remark}
	The structure of the KG algorithm can accommodate other sub-Gaussian distributions such as Bernoulli and uniform distributions. Although expressions of the acquisition function $v_{i,t}^{\rm KG}$ for these distributions could be  different, the general analysis framework in our paper can be well applied to these cases. In this research, we have assumed that the variances of the normal reward distributions are known. If we want to relax this setting to allow unknown variances, the analysis will be very difficult. In this case, the update of the posterior distribution in (\ref{bayes-update1}) and (\ref{bayes-update2}) becomes different, and there lack effective techniques to quantify the uncertainty brought by the unknown variances.
	\hfill$\blacksquare$
\end{remark}

\subsection{Alternative Analysis of PE and SR}\label{sec3.3}

PE and SR are two primary performance measures for the KG algorithm. However, the upper and lower bounds of them (Theorem \ref{thm2} and Proposition \ref{prop3}) hold only when $t>T$ for some random quantity $T$ which is not computable. Therefore, it is not clear when the bounds become valid. This is a drawback of those theoretical results.

To resolve it, a remedy is to avoid calculating $T$ and impose fixed lower bounds on the sampling rates of the arms, i.e., $\alpha_{i,n}\geq\alpha_0$ for all $i=1,2,\dots,k$, where $\alpha_0$ is a pre-specified constant with $0<\alpha_0<\frac{1}{k}$. Note that this requirement can be easily achieved by adding an initial sampling stage for the KG algorithm. 
Given $n_0\geq 2$, in the initial sampling stage, $kn_0$ rounds are separated from the budget of $n$ rounds, and each arm is pulled $n_0$ times. Then, the mean and variance of the prior distribution corresponding to each arm can be computed by use of the rewards observed in this initial sampling stage. In a similar setting, \cite{wu2018} analyzed the performance of the optimal computing budget allocation algorithms on the PE and SR. Under this setting, the PE and SR are guaranteed to exponentially converge to zero as $n\to\infty$. Below, we formally show the upper and lower bounds on the PE and SR of the KG algorithm. 

\begin{theorem}\label{thm4}
	Under the KG algorithm, the following statements hold for $\forall \alpha_0\leq\frac{n_0}{n}$:
	\begin{itemize}
		\item for PE,
		\begin{align*}
		    &\underset{j\ne b}{\min}\left\{\rule{0em}{8mm}\frac{\delta_{\rm min}}{2\sqrt{2\pi}\sigma_j\left(1+\frac{\delta_{\rm min}^2}{4\sigma_j^2}n\right)}
		    \frac{\left(\mu_b-\mu_j-\frac{\delta_{\rm min}}{2}\right)\left\lfloor\alpha_0 n\right\rfloor}{\sqrt{2\pi}\sigma_b\left(1+\frac{\left(\mu_b-\mu_j-\frac{\delta_{\rm min}}{2}\right)^2}{\sigma_b^2}n\right)}\right.\\
		    &\left.~~~~~~~\cdot\exp\left\{-\left[\frac{\delta_{\rm min}^2}{8\sigma_j^2}+\frac{\left(\mu_b-\mu_j-\frac{\delta_{\rm min}}{2}\right)^2}{2\sigma_b^2}\right]n\right\}\rule{0em}{11.5mm}\right\}\\
		    \leq&e_n\\
		    \leq&\frac{\sqrt{2}\sigma_b}{\sqrt{\pi\left\lfloor\alpha_0 n\right\rfloor}\delta_{\rm min}}\exp\left\{-\frac{\delta_{\rm min}^2}{8\sigma_b^2}\left\lfloor\alpha_0 n\right\rfloor\right\}\\
		    &+\underset{i\ne b}{\sum}\frac{\sigma_i}{\sqrt{2\pi \left\lfloor\alpha_0 n\right\rfloor}\left(\mu_b-\mu_i-\frac{\delta_{\rm min}}{2}\right)}\exp\left\{-\frac{\left(\mu_b-\mu_i-\frac{\delta_{\rm min}}{2}\right)^2}{2\sigma_i^2}\left\lfloor\alpha_0 n\right\rfloor\right\},
		\end{align*}
		\item for SR,
		\begin{align*}
		    &\underset{j\ne b}{\min}\left\{\rule{0em}{8mm}\frac{\delta_{\rm min}^2}{2\sqrt{2\pi}\sigma_j\left(1+\frac{\delta_{\rm min}^2}{4\sigma_j^2}n\right)}
		    \frac{\left(\mu_b-\mu_j-\frac{\delta_{\rm min}}{2}\right)\left\lfloor\alpha_0 n\right\rfloor}{\sqrt{2\pi}\sigma_b\left(1+\frac{\left(\mu_b-\mu_j-\frac{\delta_{\rm min}}{2}\right)^2}{\sigma_b^2}n\right)}\right.\\
		    &~~~~~~~~\left.\cdot\exp\left\{-\left[\frac{\delta_{\rm min}^2}{8\sigma_j^2}+\frac{\left(\mu_b-\mu_j-\frac{\delta_{\rm min}}{2}\right)^2}{2\sigma_b^2}\right]n\right\}\rule{0em}{12mm}\right\}\\
		    \leq&r_n\\
		    \leq&\frac{\sqrt{2}\sigma_b\delta_{\rm max}}{\sqrt{\pi\left\lfloor\alpha_0 n\right\rfloor}\delta_{\rm min}}\exp\left\{-\frac{\delta_{\rm min}^2}{8\sigma_b^2}\left\lfloor\alpha_0 n\right\rfloor\right\}\\
		    &+\underset{i\ne b}{\sum}\frac{\sigma_i\delta_{\rm max}}{\sqrt{2\pi \left\lfloor\alpha_0 n\right\rfloor}\left(\mu_b-\mu_i-\frac{\delta_{\rm min}}{2}\right)}\exp\left\{-\frac{\left(\mu_b-\mu_i-\frac{\delta_{\rm min}}{2}\right)^2}{2\sigma_i^2}\left\lfloor\alpha_0 n\right\rfloor\right\}.
		\end{align*}
	\end{itemize}
\end{theorem}

\section{Experiments}\label{sec4}

In this section, we first numerically show the convergence behavior of the KG algorithm and compare it with the bounds we derive. The test is conducted on the following two instances. 

\begin{itemize}
	\item \emph{Instance 1.} We consider a set of ten arms $\left\{1,2,\dots,10\right\}$. Set $\mu_i=1$ for $i=1,\dots,9$, $\mu_{10}=2$, and $\sigma_i=1$ for $i=1,\dots,10$. The best arm $b=10$.
	
	\item \emph{Instance 2.} We consider a set of ten arms $\left\{1,2,\dots,10\right\}$. Set $\mu_i=1$ and $\sigma_i=1$ for $i=1,\dots,5$, $\mu_i=2$ and $\sigma_i=2$ for $i=6,\dots,9$, $\mu_{10}=3$ and $\sigma_{10}=3$. The best arm $b=10$.
\end{itemize}

\begin{figure}[H]
	\vskip 0.1in
	\begin{center}
		\centerline{\includegraphics[width=0.85\linewidth]{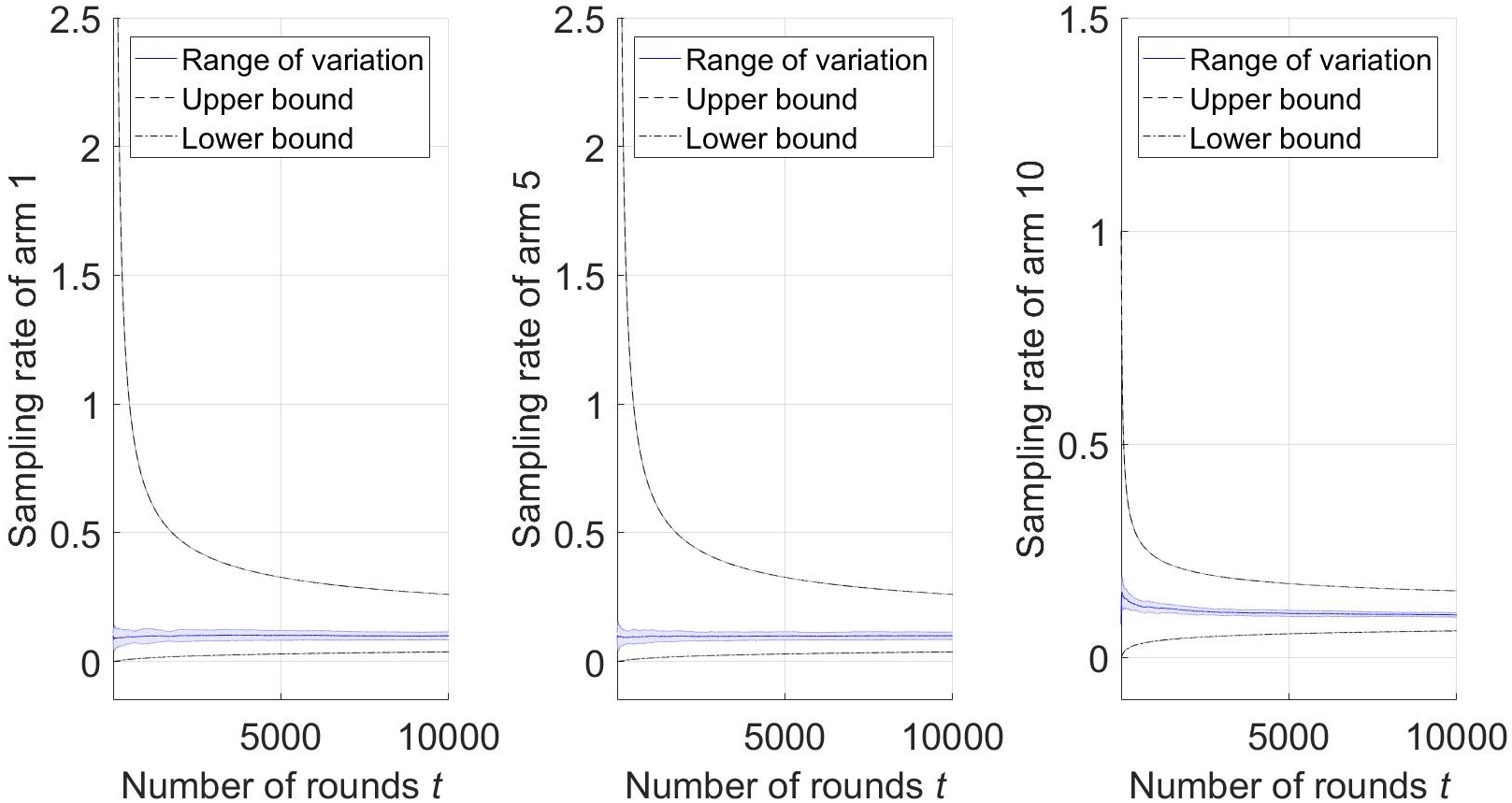}}
		\caption{Sampling rates of the three selected arms and their upper and lower bounds for instance 1.}
		\label{fig1}
	\end{center}
	\vskip -0.2in
\end{figure}

\begin{figure}[H]
	\vskip 0.1in
	\begin{center}
		\centerline{\includegraphics[width=0.85\linewidth]{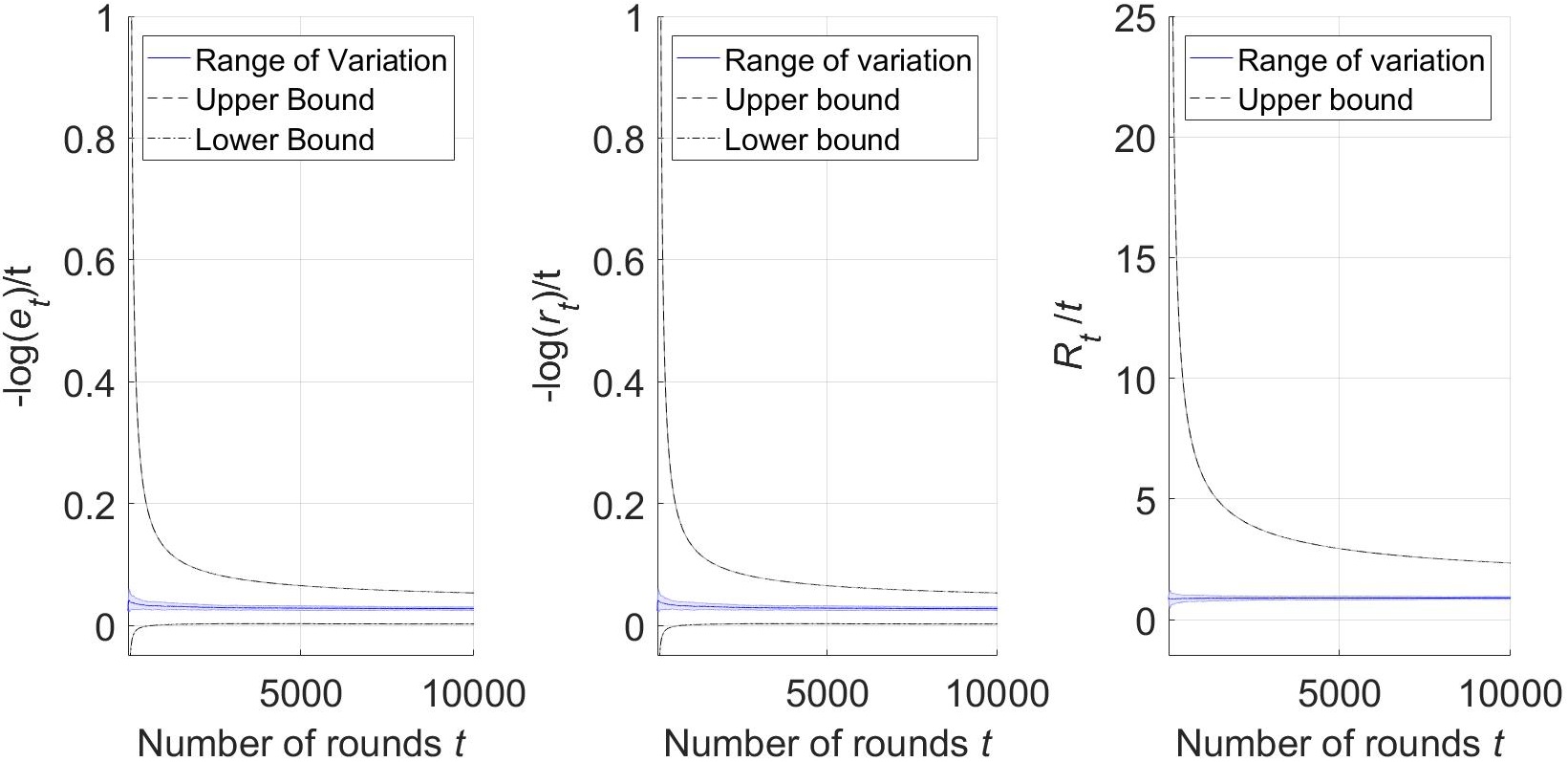}}
		\caption{PE, SR, CR, and their upper and lower bounds for instance 1.}
		\label{fig2}
	\end{center}
	\vskip -0.2in
\end{figure}

\begin{figure}[H]
	\vskip 0.1in
	\begin{center}
		\centerline{\includegraphics[width=0.85\linewidth]{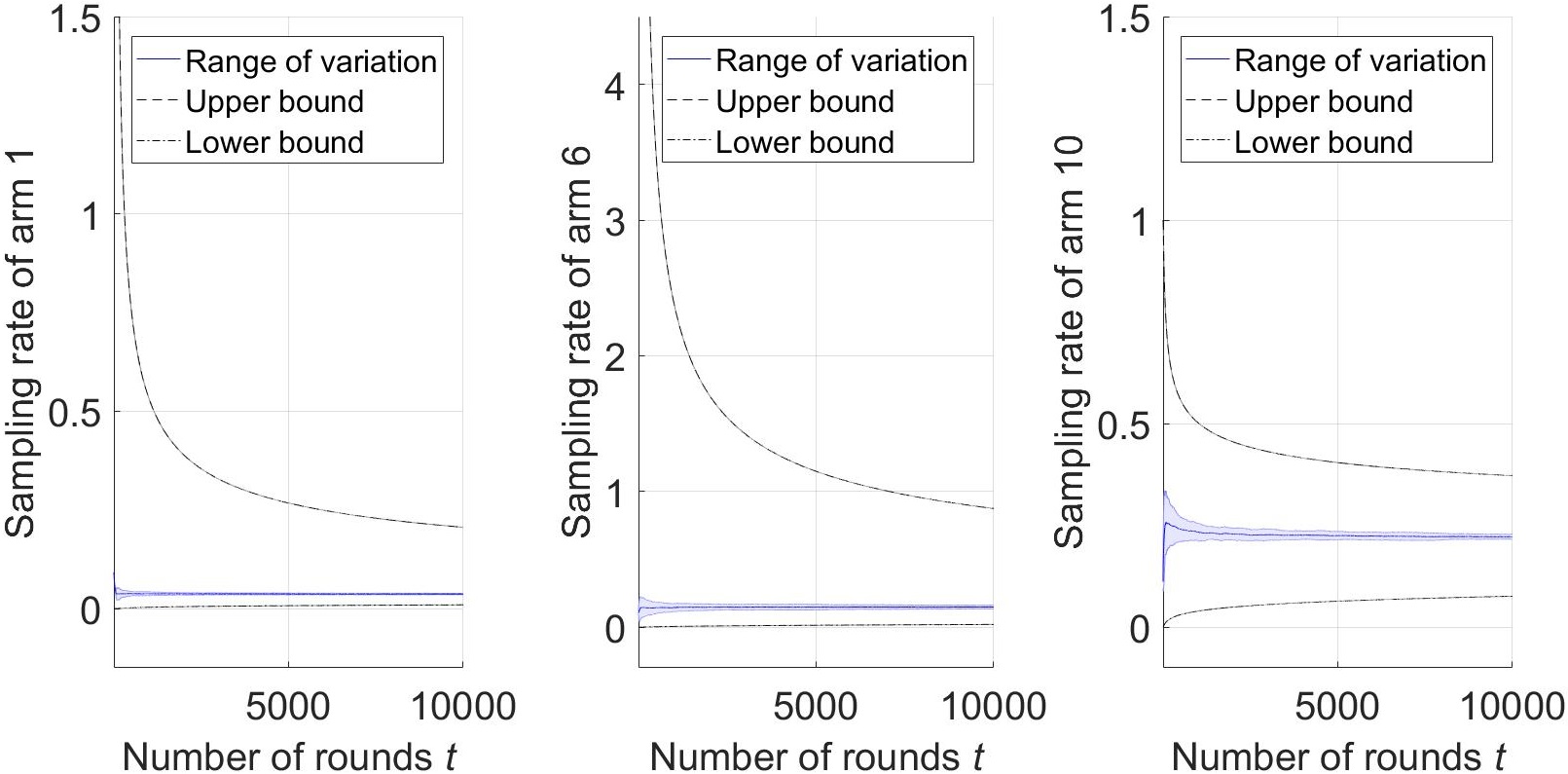}}
		\caption{Sampling rates of the three selected arms and their upper and lower bounds for instance 2.}
		\label{fig3}
	\end{center}
	\vskip -0.2in
\end{figure}

\begin{figure}[h!]
	\vskip 0.1in
	\begin{center}
		\centerline{\includegraphics[width=0.85\linewidth]{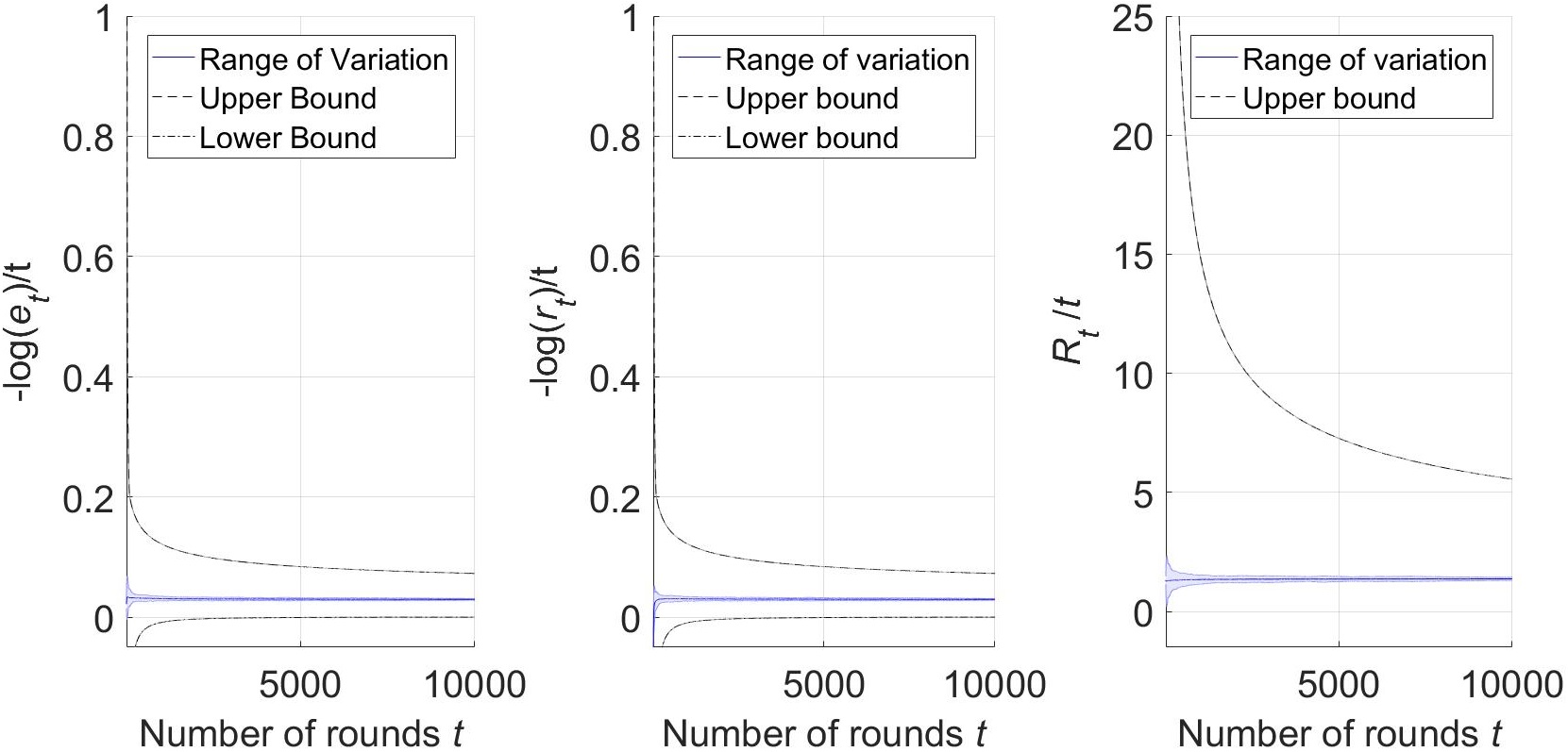}}
		\caption{PE, SR, CR, and their upper and lower bounds for instance 2.}
		\label{fig4}
	\end{center}
	\vskip -0.2in
\end{figure}

The numerical results are shown in Figures \ref{fig1}-\ref{fig4}. At the beginning, the algorithm pulls each arm for five times to obtain the initial estimates of the mean rewards of the arms. Figures \ref{fig1} and \ref{fig2} are about instance 1. Figure \ref{fig1} shows the sampling rates of three selected arms (arms 1, 5, 10) of the KG algorithm (in blue) as the number of rounds $t$ increases and the bounds of them (in black) derived in Theorem \ref{thm1}. Figure \ref{fig2} shows the three performance measures PE, SR, and CR of the algorithm (in blue), and the bounds of them (in black) derived in Theorem \ref{thm2}, Proposition \ref{prop3} and Theorem \ref{thm3}. Note that the three measures are shown on the scale of $-\frac{1}{t}\log\left(e_t\right)$, $-\frac{1}{t}\log\left(r_t\right)$ and $\frac{R_t}{t}$, and their bounds are transformed in the same way. Figures 3 and 4 show the same results for instance 2.

It is observed that the sampling rates of the selected arms and the three performance measures are well constrained by their theoretical upper bounds and lower bounds. For the sampling rates, the bounds are tighter on the best arm, and are looser on the non-best arms. For the three measures, $-\frac{1}{t}\log\left(e_t\right)$ and $-\frac{1}{t}\log\left(r_t\right)$ and their bounds have very minor difference. $\frac{R_t}{t}$ converges to a different value, but the convergence patterns of it and its bounds are similar to those of $-\frac{1}{t}\log\left(e_t\right)$ and $-\frac{1}{t}\log\left(r_t\right)$.

\begin{figure}[h!]
	\vskip 0.1in
	\begin{center}
		\centerline{\includegraphics[width=0.85\linewidth]{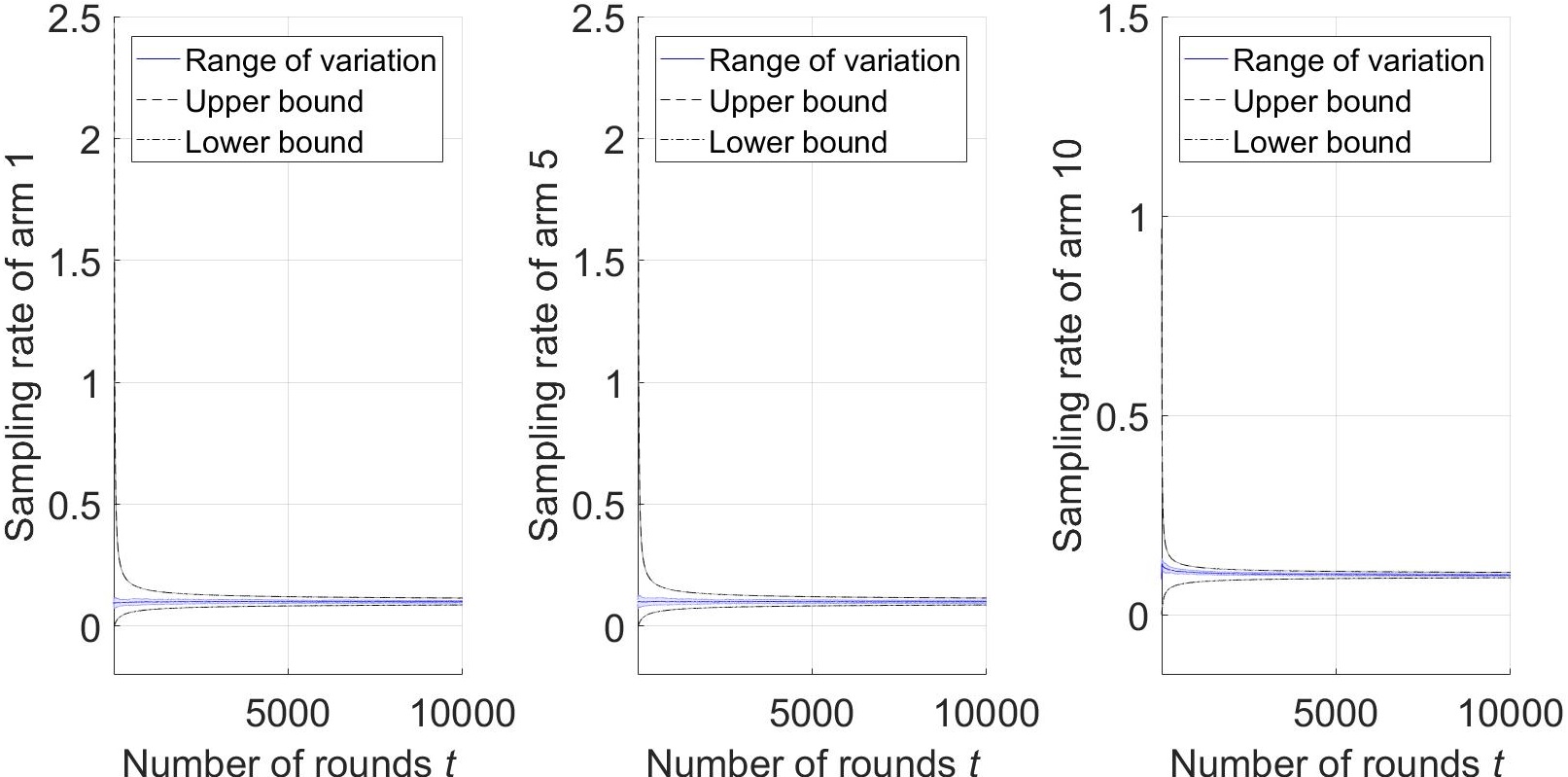}}
		\caption{Sampling rates of the three selected arms and their upper and lower bounds for instance 3.}
		\label{fig5}
	\end{center}
	\vskip -0.2in
\end{figure}

\begin{figure}[h!]
	\vskip 0.1in
	\begin{center}
		\centerline{\includegraphics[width=0.85\linewidth]{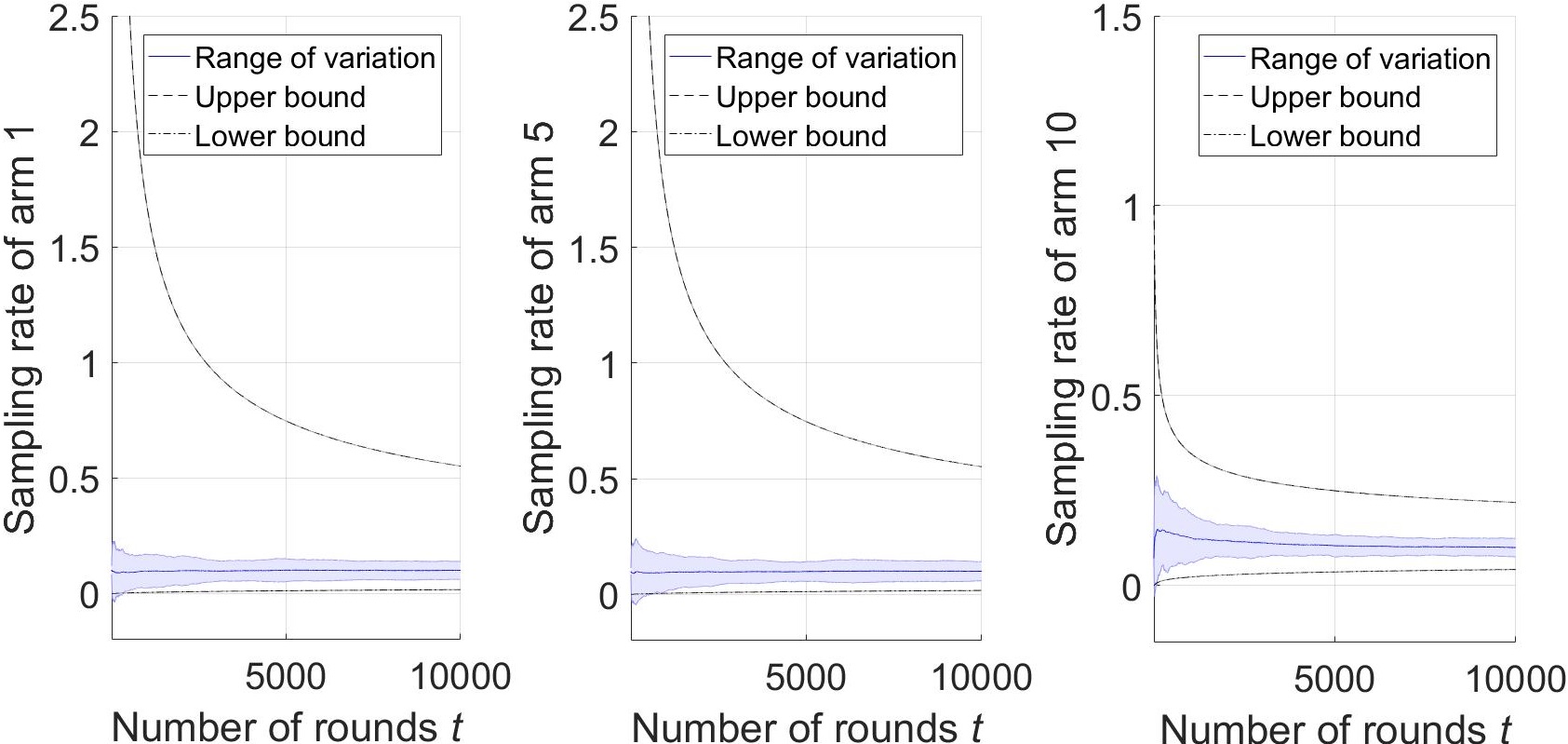}}
		\caption{Sampling rates of the three selected arms and their upper and lower bounds for instance 4.}
		\label{fig6}
	\end{center}
	\vskip -0.2in
\end{figure}

\begin{figure}[h!]
	\vskip 0.1in
	\begin{center}
		\centerline{\includegraphics[width=0.85\linewidth]{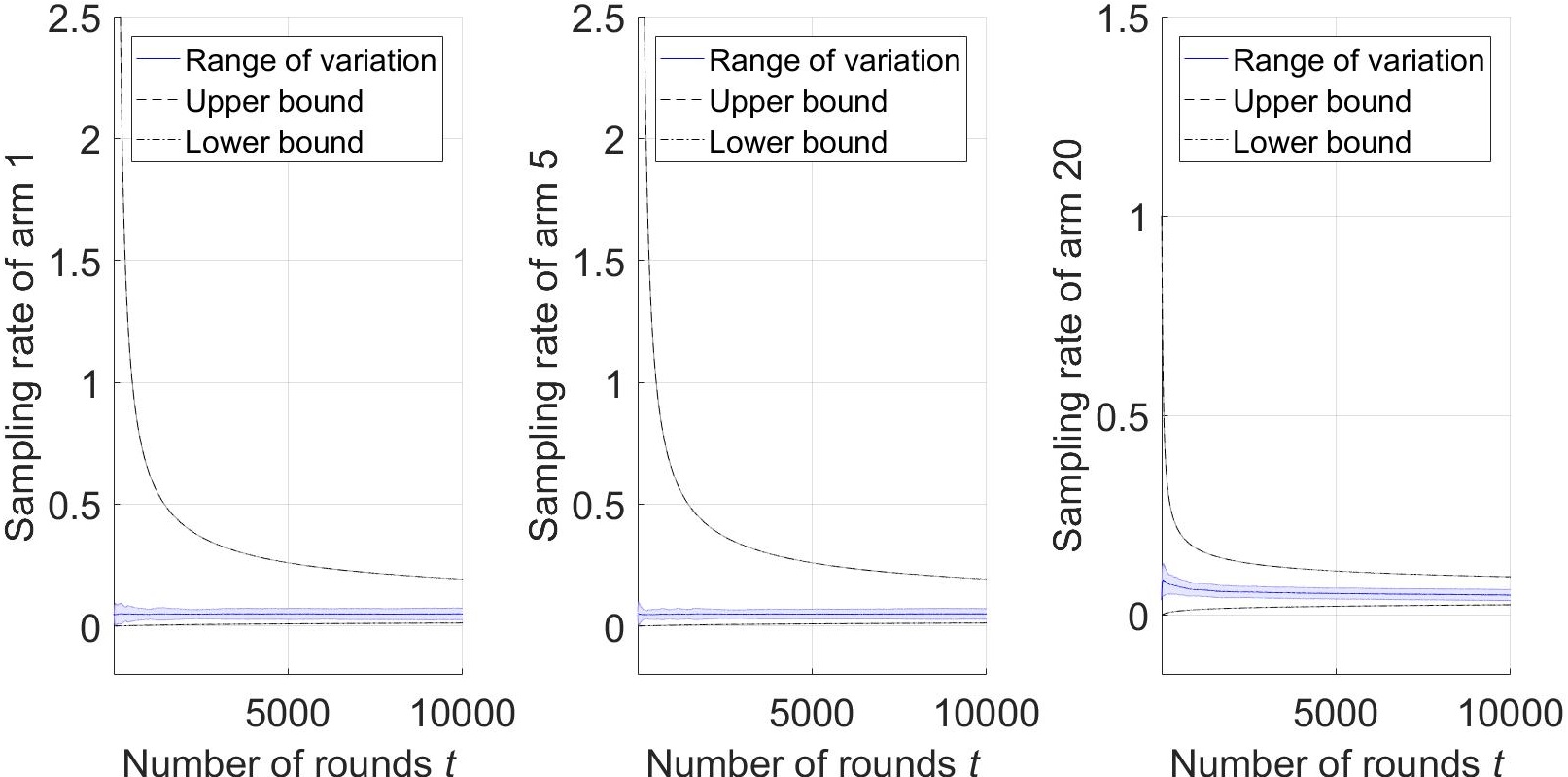}}
		\caption{Sampling rates of the three selected arms and their upper and lower bounds for instance 5.}
		\label{fig7}
	\end{center}
	\vskip -0.2in
\end{figure}

Next, we evaluate the influence of the parameters of the problem instances to the bounds of the sampling rates. The numerical test is conducted on the following three instances.
\begin{itemize}
	\item \emph{Instance 3.} We consider a set of ten arms $\left\{1,\dots,10\right\}$. Set $\mu_i=5$ for $i=1,\dots,9$, $\mu_{10}=10$, and $\sigma_i=1$ for $i=1,\dots,10$. The best arm $b=10$.
	\item \emph{Instance 4.} We consider a set of ten arms $\left\{1,\dots,10\right\}$. Set $\mu_i=1$ for $i=1,\dots,9$, $\mu_{10}=2$, and $\sigma_i=2$ for $i=1,\dots,10$. The best arm $b=10$.
	\item \emph{Instance 5.} We consider a set of twenty arms $\left\{1,\dots,20\right\}$. Set $\mu_i=1$ for $i=1,\dots,19$, $\mu_{20}=2$, and $\sigma_i=1$ for $i=1,\dots,20$. The best arm $b=20$.
\end{itemize}
The numerical results are shown in Figures \ref{fig5}-\ref{fig7}.

Comparing Figure \ref{fig5} with Figure \ref{fig1} (instances 3 and 1), we can see that in Figure \ref{fig5}, the ranges of variations are narrower, and the upper and lower bounds of the sampling rates are tighter. Since instance 3 has a larger gap in means between the arms than instance 1, it suggests that increasing this gap tends to tighten the bounds of the sampling rates. Comparing Figure \ref{fig6} with Figure \ref{fig1} (instances 4 and 1), we can see that in Figure \ref{fig6}, the ranges of variations are slightly wider, and the upper and lower bounds of the sampling rates are looser. Since in instance 4, variances of each arm are larger than those in instance 1, it suggests that increasing the variances tends to loosen the bounds of the sampling rates. Comparing Figure \ref{fig7} with Figure \ref{fig1} (instances 5 and 1), we can see that the upper and lower bounds of the sampling rates are both smaller in Figure \ref{fig7} than in Figure \ref{fig1}, and the bounds are slightly tighter in Figure \ref{fig7}. Since instance 5 has more arms than instance 1, it suggests that increasing the number of arms tends to tighten the bounds of the sampling rates.

\section{Conclusions and Discussion}\label{sec5}

The KG algorithm is a popular and effective algorithm for the BAI problem, but existing theoretical treatment to it is mostly limited to its asymptotic characteristics. In this paper, we explore the finite-time performance of the KG algorithm. We consider the measures of the probability of error and simple regret in the BAI problem and the measure of cumulative regret in the MAB problem, and derive bounds of these measures. At last, these bounds are illustrated using numerical examples.

Our analysis can serve as the ground for future research on the KG algorithm and BAI. In the literature, the KG algorithm has been extended to solve other types of sequential decision problems, such as the multi-objective MAB \cite{yahyaa2014}, parallel BAI \cite{wu2016}, and contextual bandits \cite{ding2021}. In addition, there are some other BAI algorithms that share similar mechanisms and structures as KG, e.g., the probability improvement \cite{kashner1964}, expected improvement \cite{jones1998}, etc. The analysis in this research might be extended to study the performance of these KG-type and improvement-based algorithms. In addition, the validity of the bounds of PE, SR and CR in Section \ref{sec3.2} highly depends on quantity $T$ which is difficult to compute. Therefore, it is an important future research direction to study how to quantify $T$.


\bibliography{FTP_KG_Ref}
\bibliographystyle{plain}

\newpage
\appendix
\onecolumn

\renewcommand\thesection{A.\arabic{section}}
\setcounter{equation}{0}
\renewcommand\theequation{A.\arabic{equation}}

\section{Proof of Lemma \ref{lemma2}}

Notice that $\hat{Y}_m\sim\mathcal{N}\left(\mu,\frac{\sigma^{2}}{m}\right)$. Let $Z=\frac{\sqrt{m}\left(\hat{Y}_m-\mu\right)}{\sigma}\sim\mathcal{N}\left(0,1\right)$. For the probability density function $\phi(x)$ of the standard normal distribution and $\forall\epsilon\geq 0$,
\begin{align*}
\mathbb{P}\left(Z\geq\frac{\sqrt{m}\epsilon}{\sigma}\right)\leq&\frac{\sigma}{\sqrt{m}\epsilon}\int_{\frac{\sqrt{m}\epsilon}{\sigma}}^{+\infty}z\phi(z)\text{d}z
=-\frac{\sigma}{\sqrt{m}\epsilon}\int_{\frac{\sqrt{m}\epsilon}{\sigma}}^{+\infty}\phi'(z)\text{d}z=\frac{\sigma}{\sqrt{m}\epsilon}\phi\left(\frac{\sqrt{m}\epsilon}{\sigma}\right)\\
\leq&\frac{\sigma}{\sqrt{m}\epsilon}\exp\left\{-\frac{m\epsilon^{2}}{2\sigma^{2}}\right\}.
\end{align*}
By symmetry, it holds that 
\begin{align*}
\mathbb{P}\left(\left|Z\right|\geq\frac{\sqrt{m}\epsilon}{\sigma}\right)\leq\frac{2\sigma}{\sqrt{m}\epsilon}\exp\left\{-\frac{m\epsilon^{2}}{2\sigma^{2}}\right\},
\end{align*}
i.e., $\mathbb{P}\left(\left|\hat{Y}_m-\mu\right|\geq\epsilon\right)\leq\frac{2\sigma}{\sqrt{m}\epsilon}\exp\left\{-\frac{m\epsilon^{2}}{2\sigma^{2}}\right\}$.
\hfill $\blacksquare$

\section{Proof of Lemma \ref{lemma3}}

Notice that $f(-x)>\frac{\phi(x)}{x^{3}}$ holds based on Lemma 4 of \cite{qin2017}. According to \cite{small2010},
\begin{align*}
\frac{1-\Phi(x)}{\phi(x)}>\frac{1}{x}-\frac{1}{x^{3}}
\end{align*} 
for $\forall x>0$. Then, 
\begin{align*}
f(-x)<\phi(x)-x\left(\frac{1}{x}-\frac{1}{x^{3}}\right)\phi(x)=\frac{\phi(x)}{x^{2}}.
\end{align*} 
\hfill $\blacksquare$

\section{Proof of Proposition \ref{prop1}}

Denote by  $A_{t,\tau}=\left\{i\left|N_{i,t}<\tau^{\frac{7}{8}}\right.\right\}$, $B_{t,\tau}=\left\{i\left|N_{i,t}<\tau^{\frac{3}{4}}\right.\right\}$, $\bar{A}_{t,\tau}=\left\{1,\dots,k\right\}\setminus A_{t,\tau}$, $\bar{B}_{t,\tau}=\left\{1,\dots,k\right\}\setminus B_{t,\tau}$, $\forall t\in\mathbb{N}$, $\forall\tau\in\mathbb{R}_+$. Without loss of generality, we assume that $b=1$ throughout the proof. Let $\hat{\jmath}_t$ denote the arm with the $j$-th largest estimated mean in round $t$, $j\in\left\{1,\dots,k\right\}$, $\forall t\in\mathbb{N}$. 

We first consider the scenario of $k\geq 4$. The proof is divided into three stages:

\begin{itemize}
	\item[1.] Prove that for $\tau_1=\max\left\{\left(\frac{2\sqrt{2}\sigma_{\rm max}\left(\delta_{\rm max}+2\sigma_{\rm max}W\right)}{\sigma_{\rm min}\delta_{\rm min}}\right)^{8},\left(\frac{4\sigma_{\rm max}W}{\delta_{\rm min}}\right)^{8}\right\}$, $\forall\tau\geq\tau_1$, $\forall t\leq k\tau$, if $B_{t,\tau}\ne\emptyset$, then $I_t\in A_{t,\tau}$. Equivalently, we prove by contradiction that for $\forall\tau\geq\tau_1$, $\forall t\leq k\tau$, if $I_t\in\bar{A}_{t,\tau}$, then $B_{t,\tau}=\emptyset$. To this end, we first prove by contradiction that $\hat{\jmath}_{t}\in B_{t,\tau}$ in an ascending order of $j$.
	
	\begin{itemize}
		\item[(1)] Suppose that $\exists\tau_0>\tau_1$, $\exists t_0\leq k\tau_0$, $\hat{1}_{t_0}\in B_{t_0,\tau_0}$.
		
		\begin{itemize}
			\item[(1.1)] If $I_{t_0}=\hat{1}_{t_0}$, then $\hat{1}_{t_0}\in \bar{A}_{t_0,\tau_0}$. It contradicts that $\hat{1}_{t_0}\in B_{t_0,\tau_0}$. So $\hat{1}_{t_0}\notin B_{t_0,\tau_0}$ in this case.
			
			\item[(1.2)] If $I_{t_0}\ne\hat{1}_{t_0}$, for $\tau_0>\tau_1$,
			\begin{align*}
			v_{\hat{1}_{t_0},t_0}^{\text{KG}}
			>&\frac{\sigma_{\hat{1}_{t_0}}}{\sqrt{\tau_0^{\frac{3}{4}}\left(\tau_0^{\frac{3}{4}}+1\right)}}f\left(\frac{\theta_{\hat{2}_{t_0},t_0}-\theta_{\hat{1}_{t_0},t_0}}{\sigma_{\hat{1}_{t_0}}}\sqrt{\tau_0^{\frac{3}{4}}\left(\tau_0^{\frac{3}{4}}+1\right)}\right)\\
			>&\frac{\sigma_{\hat{1}_{t_0}}}{\sqrt{2}\tau_0^{\frac{3}{4}}}f\left(\frac{\theta_{\hat{2}_{t_0},t_0}-\theta_{\hat{1}_{t_0},t_0}}{\sigma_{\hat{1}_{t_0}}}\sqrt{2}\tau_0^{\frac{3}{4}}\right)
			>\frac{\sigma_{I_{t_0}}}{\tau_0^{\frac{7}{8}}}f\left(\frac{\theta_{I_{t_0},t_0}-\theta_{\hat{1}_{t_0},t_0}}{\sigma_{I_{t_0}}}\tau_0^{\frac{7}{8}}\right)\\
			>&\frac{\sigma_{I_{t_0}}}{\sqrt{\tau_0^{\frac{7}{8}}\left(\tau_0^{\frac{7}{8}}+1\right)}}f\left(\frac{\theta_{I_{t_0},t_0}-\theta_{\hat{1}_{t_0},t_0}}{\sigma_{I_{t_0}}}\sqrt{\tau_0^{\frac{7}{8}}\left(\tau_0^{\frac{7}{8}}+1\right)}\right)
			\geq v_{I_{t_0},t_0}^{\text{KG}}.
			\end{align*}
			The first and second inequalities hold because $f(x)$ is monotone increasing with respect to $x$ and $N_{\hat{1}_{t_0},t_0}<\tau_0^{\frac{3}{4}}$. The third inequality holds because $\tau_0>\frac{16\sigma_{\rm max}^{8}}{\sigma_{\rm min}^{8}}$ and $\theta_{\hat{2}_{t_0},t_0}\geq\theta_{I_{t_0},t_0}$. The last inequality holds because $f(x)$ is monotone increasing with respect to $x$ and $N_{I_{t_0},t_0}\geq\tau_0^{\frac{7}{8}}$. Notice that $v_{\hat{1}_{t_0},t_0}^{\text{KG}}>v_{I_{t_0},t_0}^{\text{KG}}$ contradicts the definition of $I_{t_0}$. So $\hat{1}_{t_0}\in\bar{B}_{t_0,\tau_0}$. 
			
		\end{itemize}
		
		\item[(2)] Suppose that $\exists\tau_0>\tau_1$, $\exists t_0\leq k\tau_0$, $\hat{2}_{t_0}\in B_{t_0,\tau_0}$.
		
		\begin{itemize}
			\item[(2.1)] If $I_{t_0}=\hat{2}_{t_0}$, then $\hat{2}_{t_0}\in \bar{A}_{t_0,\tau_0}$. It contradicts that $\hat{2}_{t_0}\in B_{t_0,\tau_0}$. So $\hat{2}_{t_0}\notin B_{t_0,\tau_0}$ in this case.
			
			\item[(2.2)] If $I_{t_0}=\hat{1}_{t_0}$, for $\tau_0>\tau_1$,
			\begin{align*}
			v_{\hat{2}_{t_0},t_0}^{\text{KG}}
			>\frac{\sigma_{\hat{2}_{t_0}}}{\sqrt{2}\tau_0^{\frac{3}{4}}}f\left(\frac{\theta_{\hat{2}_{t_0},t_0}-\theta_{\hat{1}_{t_0},t_0}}{\sigma_{\hat{2}_{t_0}}}\sqrt{2}\tau_0^{\frac{3}{4}}\right)
			>\frac{\sigma_{\hat{1}_{t_0}}}{\tau_0^{\frac{7}{8}}}f\left(\frac{\theta_{\hat{2}_{t_0},t_0}-\theta_{\hat{1}_{t_0},t_0}}{\sigma_{\hat{1}_{t_0}}}\tau_0^{\frac{7}{8}}\right)
			>v_{\hat{1}_{t_0},t_0}^{\text{KG}}.
			\end{align*}
			It contradicts the definition of $I_{t_0}=\hat{1}_{t_0}$. So $\hat{2}_{t_0}\notin B_{t_0,\tau_0}$ in this case.
			
			\item[(2.3)] If $I_{t_0}\ne\hat{1}_{t_0},\hat{2}_{t_0}$, for $\tau_0>\tau_1$, following similar discussion as in Case (1.2),
			\begin{align*}
			v_{\hat{2}_{t_0},t_0}^{\text{KG}}
			>&\frac{\sigma_{\hat{2}_{t_0}}}{\sqrt{2}\tau_0^{\frac{3}{4}}}f\left(\frac{\theta_{I_{t_0},t_0}-\theta_{\hat{1}_{t_0},t_0}}{\sigma_{\hat{2}_{t_0}}}\sqrt{2}\tau_0^{\frac{3}{4}}\right)\\
			>&\frac{\sigma_{I_{t_0}}}{\sqrt{\tau_0^{\frac{7}{8}}\left(\tau_0^{\frac{7}{8}}+1\right)}}f\left(\frac{\theta_{I_{t_0},t_0}-\theta_{\hat{1}_{t_0},t_0}}{\sigma_{I_{t_0}}}\sqrt{\tau_0^{\frac{7}{8}}\left(\tau_0^{\frac{7}{8}}+1\right)}\right)
			\geq v_{I_{t_0},t_0}^{\text{KG}}.
			\end{align*}
			It contradicts the definition of $I_{t_0}$. So $\hat{2}_{t_0}\in\bar{B}_{t_0,\tau_0}$.
			
		\end{itemize} 
		
		\item[(3)] Suppose that $\exists\tau_0>\tau_1$, $\exists t_0\leq k\tau_0$, $\hat{\jmath}_{t_0}\in B_{t_0,\tau_0}$ where $j\in\left\{3,\dots,k-1\right\}$.
		
		\begin{itemize}
			\item[(3.1)] If $I_{t_0}=\hat{\jmath}_{t_0}$, then $\hat{\jmath}_{t_0}\in \bar{A}_{t_0,\tau_0}$. It contradicts that $\hat{\jmath}_{t_0}\in B_{t_0,\tau_0}$. 
			
			\item[(3.2)] If $I_{t_0}=\hat{1}_{t_0}$, for $\tau_0>\tau_1$,
			\begin{align*}
			v_{\hat{\jmath}_{t_0},t_0}^{\text{KG}}
			>&\frac{\sigma_{\hat{\jmath}_{t_0}}}{\sqrt{\tau_0^{\frac{3}{4}}\left(\tau_0^{\frac{3}{4}}+1\right)}}f\left(\frac{\theta_{\hat{\jmath}_{t_0},t_0}-\theta_{\hat{1}_{t_0},t_0}}{\sigma_{\hat{\jmath}_{t_0}}}\sqrt{\tau_0^{\frac{3}{4}}\left(\tau_0^{\frac{3}{4}}+1\right)}\right)\\
			>&\frac{\sigma_{\hat{\jmath}_{t_0}}}{\sqrt{2}\tau_0^{\frac{3}{4}}}f\left(\frac{2\delta_{\rm max}+4\sigma_{\rm max}W}{\delta_{\rm min}}\frac{\theta_{\hat{2}_{t_0},t_0}-\theta_{\hat{1}_{t_0},t_0}}{\sigma_{\hat{\jmath}_{t_0}}}\sqrt{2}\tau_0^{\frac{3}{4}}\right)\\
			>&\frac{\sigma_{\hat{1}_{t_0}}}{\tau_0^{\frac{7}{8}}}f\left(\frac{\theta_{\hat{2}_{t_0},t_0}-\theta_{\hat{1}_{t_0},t_0}}{\sigma_{\hat{1}_{t_0}}}\tau_0^{\frac{7}{8}}\right)\\
			>&\frac{\sigma_{\hat{1}_{t_0}}}{\sqrt{\tau_0^{\frac{7}{8}}\left(\tau_0^{\frac{7}{8}}+1\right)}}f\left(\frac{\theta_{\hat{2}_{t_0},t_0}-\theta_{\hat{1}_{t_0},t_0}}{\sigma_{\hat{1}_{t_0}}}\sqrt{\tau_0^{\frac{7}{8}}\left(\tau_0^{\frac{7}{8}}+1\right)}\right)
			\geq v_{\hat{1}_{t_0},t_0}^{\text{KG}}.
			\end{align*}
			The second inequality holds because $\left|\theta_{\hat{\jmath}_{t_0},t_0}-\theta_{\hat{1}_{t_0},t_0}\right|<\delta_{\rm max}+2\sigma_{\rm max}W$ and $\left|\theta_{\hat{2}_{t_0},t_0}-\theta_{\hat{1}_{t_0},t_0}\right|>\frac{\delta_{\rm min}}{2}$ when $\tau_0>\max\left\{16,\left(\frac{4\sigma_{\rm max}W}{\delta_{\rm min}}\right)^{8}\right\}$, which results from Lemma \ref{lemma1}. The third inequality holds because $f(x)$ is monotone increasing with respect to $x$ and $\tau_0>\left(\frac{2\sqrt{2}\sigma_{\rm max}\left(\delta_{\rm max}+2\sigma_{\rm max}W\right)}{\sigma_{\rm min}\delta_{\rm min}}\right)^{8}$. Notice that $v_{\hat{1}_{t_0},t_0}^{\text{KG}}>v_{I_{t_0},t_0}^{\text{KG}}$ contradicts the definition of $I_{t_0}$. So $\hat{\jmath}_{t_0}\notin B_{t_0,\tau_0}$ in this case.
			
			\item[(3.3)] If $I_{t_0}\in\left\{\left.\widehat{j'}_{t_0}\right|j'=2,\dots,j-1\right\}$, for $\tau_0>\tau_1$, following similar discussion as in Case (3.2),
			\begin{align*}
			v_{\hat{\jmath}_{t_0},t_0}^{\text{KG}}
			>&\frac{\sigma_{\hat{\jmath}_{t_0}}}{\sqrt{2}\tau_0^{\frac{3}{4}}}f\left(\frac{2\delta_{\rm max}+4\sigma_{\rm max}W}{\delta_{\rm min}}\frac{\theta_{I_t,t_0}-\theta_{\hat{1}_{t_0},t}}{\sigma_{\hat{2}_{t_0}}}\sqrt{2}\tau_0^{\frac{3}{4}}\right)\\
			>&\frac{\sigma_{I_{t_0}}}{\sqrt{\tau_0^{\frac{7}{8}}\left(\tau_0^{\frac{7}{8}}+1\right)}}f\left(\frac{\theta_{I_{t_0},t_0}-\theta_{\hat{1}_{t_0},t_0}}{\sigma_{I_{t_0}}}\sqrt{\tau_0^{\frac{7}{8}}\left(\tau_0^{\frac{7}{8}}+1\right)}\right)
			\geq v_{I_{t_0},t_0}^{\text{KG}}.
			\end{align*}
			It contradicts the definition of $I_{t_0}$. So $\hat{\jmath}_{t_0}\notin B_{t_0,\tau_0}$ in this case.
			
			\item[(3.4)] If $I_{t_0}\in\left\{\left.\widehat{j'}_{t_0}\right|j'=j+1,\dots,k\right\}$, for $\tau_0>\tau_1$, following similar discussion as in Case (1.2),
			\begin{align*}
			v_{\hat{k}_{t_0},t_0}^{\text{KG}}
			>&\frac{\sigma_{\hat{k}_{t_0}}}{\sqrt{2}\tau_0^{\frac{3}{4}}}f\left(\frac{\theta_{I_{t_0},t_0}-\theta_{\hat{1}_{t_0},t_0}}{\sigma_{\hat{k}_{t_0}}}\sqrt{2}\tau_0^{\frac{3}{4}}\right)\\
			>&\frac{\sigma_{I_{t_0}}}{\sqrt{\tau_0^{\frac{7}{8}}\left(\tau_0^{\frac{7}{8}}+1\right)}}f\left(\frac{\theta_{I_{t_0},t_0}-\theta_{\hat{1}_{t_0},t_0}}{\sigma_{I_{t_0}}}\sqrt{\tau_0^{\frac{7}{8}}\left(\tau_0^{\frac{7}{8}}+1\right)}\right)
			\geq v_{I_{t_0},t_0}^{\text{KG}}.
			\end{align*}
			It contradicts the definition of $I_{t_0}$. So $\hat{\jmath}_{t_0}\in\bar{B}_{t_0,\tau_0}$. 
			
		\end{itemize}
		
		\item[(4)] Suppose that $\exists\tau_0>\tau_1$, $\exists t_0\leq k\tau_0$, $\hat{k}_{t_0}\in B_{t_0,\tau_0}$.
		
		\begin{itemize}
			\item[(4.1)] If $I_{t_0}=\hat{k}_{t_0}$, then $\hat{k}_{t_0}\in\bar{A}_{t_0,\tau_0}$. It contradicts that $\hat{k}_{t_0}\in B_{t_0,\tau_0}$. 
			
			\item[(4.2)] If $I_{t_0}\ne\hat{k}_{t_0}$, following similar discussion as in Cases (3.2) and (3.3), we can prove that $\hat{k}_{t_0}\in B_{t_0,\tau_0}$ leads to contradiction to the definition of $I_{t_0}$. So $\hat{k}_{t_0}\in\bar{B}_{t_0,\tau_0}$.
			
		\end{itemize}
		
		Thus, $\forall\tau\geq\tau_1$, $\forall t\leq k\tau$, if $I_t\in\bar{A}_{t,\tau}$, then $B_{t,\tau}=\emptyset$. That is, $\forall\tau\geq\tau_1$, $\forall t\leq k\tau$, if $B_{t,\tau}\ne\emptyset$, then $I_t\in A_{t,\tau}$. In the scenario of $k=2$, we can follow similar discussion as in Cases (1) and (2) to show that $\forall\tau\geq\tau_1$, $\forall t\leq k\tau$, if $B_{t,\tau}\ne\emptyset$, then $I_t\in A_{t,\tau}$. In the scenario of $k=3$, we can follow similar discussion as in Cases (1), (2) and (4) to show that $\forall\tau\geq\tau_1$, $\forall t\leq k\tau$, if $B_{t,\tau}\ne\emptyset$, then $I_t\in A_{t,\tau}$. 
		
	\end{itemize}
	
	\item[2.] Prove by contradiction that for $\tau_2=\max\left\{256k^{8},\tau_1\right\}$, $\forall\tau>\tau_2$, $B_{\left\lfloor k\tau\right\rfloor,\tau}=\emptyset$, where $\left\lfloor x\right\rfloor$ denotes the largest integer no greater than $x$. Suppose that $\exists\tau_0>\tau_2$, $B_{\left\lfloor k\tau_0\right\rfloor,\tau_0}\ne\emptyset$. Then, $A_{t,\tau_0}\ne\emptyset$ and $B_{t,\tau_0}\ne\emptyset$ for $t=1,\dots,\left\lfloor k\tau_0\right\rfloor$. Notice that $\left\lfloor\tau_0\right\rfloor-1\geq k\tau_0^{\frac{7}{8}}$ for $\tau_0>\tau_2$. It implies that at least one arm is pulled at least $\tau_0^{\frac{7}{8}}$ times before round $\left\lfloor\tau_0\right\rfloor$. That is, $\left|A_{\left\lfloor\tau_0\right\rfloor,\tau_0}\right|\leq k-1$, where $\left|S\right|$ for the set $S$ denotes the number of elements in the set $S$.  For $\forall j\in\left\{2,\dots,k\right\}$, $\forall l\in\left\{\left\lfloor\left(j-1\right)\tau_0\right\rfloor,\dots,\left\lfloor j\tau_0\right\rfloor-1\right\}$, $B_{l,\tau_0}\ne\emptyset$. Following similar discussion as in Stage 1, we can prove that $I_t\in A_{t,\tau_0}$ for $\forall l\in\left\{\left\lfloor\left(j-1\right)\tau_0\right\rfloor,\dots,\left\lfloor j\tau_0\right\rfloor-1\right\}$, $\tau_0>\tau_1$, and thus $\sum_{i\in A_{\left\lfloor\left(j-1\right)\tau_0\right\rfloor,\tau_0}}\left(N_{i,\left\lfloor j\tau_0\right\rfloor}-N_{i,\left\lfloor \left(j-1\right)\tau_0\right\rfloor}\right)\geq\left\lfloor \tau_0\right\rfloor-1\geq k\tau_0^{\frac{7}{8}}$. It indicates that at least one arm in $A_{\left\lfloor\left(j-1\right)\tau_0\right\rfloor,\tau_0}$ is pulled at least $\tau_0^{\frac{7}{8}}$ times during rounds $\left\lfloor\left(j-1\right)\tau_0\right\rfloor,\cdots,\left\lfloor j\tau_0\right\rfloor-1$, $j\in\left\{2,\dots,k\right\}$. Then, $A_{\left\lfloor k\tau_0\right\rfloor,\tau_0}=\emptyset$. It contradicts that $A_{t,\tau_0}\ne\emptyset$ for $t=\left\lfloor k\tau_0\right\rfloor$, which implies that $\forall\tau>\tau_2$, $B_{\left\lfloor k\tau\right\rfloor,\tau}=\emptyset$.
	
	\item[3.] Let $\tau=\frac{t}{k}$. We have $B_{t,\frac{t}{k}}=\left\{i\left|N_{i,t}<\left(\frac{t}{k}\right)^{\frac{3}{4}}\right.\right\}$. Following similar discussion as in Stage 2, we can prove that for $\forall t>k\tau_2$, $B_{t,\frac{t}{k}}=\emptyset$, that is, $N_{i,t}\geq\left(\frac{t}{k}\right)^{\frac{3}{4}}$ for $\forall i\in\left\{1,\dots,k\right\}$.
	
\end{itemize}
\hfill$\blacksquare$

\section{Proof of Proposition \ref{prop2}}

Recall that for $\forall t$, $\forall i$, $\hat{\mu}_{i,t}=\frac{1}{N_{i,t}}\sum_{s=1}^{t}\mathbbm{1}\left\{I_s=i\right\}X_{i,s}$ denotes the estimated mean reward of arm $i$ after round $t$. Under the frequentist setting, we treat $\hat{\mu}_{i,t}$ as a normal random variable. According to Lemma \ref{lemma2} and Proposition \ref{prop1}, for  $\forall t>T_0$, $\forall i$,
\begin{align*}
\mathbb{P}\left(\left|\hat{\mu}_{i,s}-\mu_i\right|\geq\frac{\sqrt{k} s^{-\frac{1}{4}}}{2}\right)
\leq 4\sigma_{\rm max} k^{-\frac{1}{8}}s^{-\frac{1}{8}}\exp\left\{-\frac{k^{\frac{1}{4}}s^{\frac{1}{4}}}{8\sigma_{\max}^{2}}\right\}\triangleq q(s).
\end{align*}
Then, $\exists T_1=\max\left\{T_0,\frac{16k^2}{\delta_{\rm min}^{4}}\right\}$, for $\forall t>T_1$, with a probability of at least $1-q(t)$, $\left|\hat{\mu}_{i,t}-\mu_i\right|<\frac{\delta_{\rm min}}{2}$ for $\forall i$, that is, the true best arm $b$ can be correctly selected under the KG algorithm. In addition, $\exists T_2=\max\left\{T_1,16\delta_{\rm min}^{-4},k\left(\frac{4\sigma_{\rm max}}{\delta_{\rm min}}\right)^{\frac{4}{3}}\right\}$, for $\forall i\ne b$, $\forall t>T_2$, with a probability of at least $1-q(t)$, $\left|\hat{\mu}_{b,t}-\hat{\mu}_{i,t}\right|>\frac{\delta_{\rm min}}{2}$, $\frac{\left|\hat{\mu}_{b,t}-\hat{\mu}_{i,t}\right|}{\sigma_i}\sqrt{N_{i,t}\left(N_{i,t}+1\right)}>2$.

In the analysis below, we replace $\theta_{i,t}$ in (\ref{value-kg}) by $\hat{\mu}_{i,t}$, $\forall i$, $\forall t$. According to Lemma \ref{lemma3}, $\exists T_3=\max\left\{T_2,k\left(\frac{4\sigma_{\rm max}^{3}}{\delta_{\rm min}^{2}}\right)^{\frac{4}{9}}\right\}$, for $\forall i\ne b$, $\forall t>T_3$, with a probability of at least $\left[1-q(t)\right]^{k}$,
\begin{align}
v_{i,t}^{\rm KG}
>&\frac{\sigma_i^{4}}{\left(\hat{\mu}_{b,t}-\hat{\mu}_{i,t}\right)^{3}}N_{i,t}^{-2}\left(N_{i,t}+1\right)^{-2}
\phi\left(\frac{\hat{\mu}_{i,t}-\hat{\mu}_{b,t}}{\sigma_i}\sqrt{N_{i,t}\left(N_{i,t}+1\right)}\right)\nonumber\\
>&\frac{1}{\sqrt{2\pi}}\exp\left\{-\frac{\left(\mu_b-\mu_i+t^{-\frac{1}{4}}\right)^{2}}{2\sigma_i^{2}}\left(N_{i,t}+1\right)^{2}+\ln\left(\frac{\sigma_i^{4}}{\left(\mu_b-\mu_i+t^{-\frac{1}{4}}\right)^{3}\left(N_{i,t}+1\right)^{4}}\right)\right\}\nonumber\\
>&\frac{1}{\sqrt{2\pi}}\exp\left\{-\frac{\left(\mu_b-\mu_i+t^{-\frac{1}{4}}\right)^{2}}{2\sigma_i^{2}}\left(N_{i,t}+1\right)^{2}+\ln\left(\frac{8\sigma_{\rm min}^{4}}{27\delta_{\rm max}^{3}\left(t+1\right)^{4}}\right)\right\}\nonumber\\
>&\frac{1}{\sqrt{2\pi}}\exp\left\{-\left[\frac{\left(\mu_b-\mu_i+t^{-\frac{1}{4}}\right)^{2}}{2\sigma_i^{2}}+\frac{4k}{\sqrt{t}}+\frac{k\max\left\{\ln\left(\frac{27\delta_{\rm max}^{3}}{8\sigma_{\rm min}^{4}}\right),0\right\}}{t}\right]\left(N_{i,t}+1\right)^{2}\right\}
\triangleq\underline{v}_{i,t}^{\rm KG},\label{value-kg-nb-low}
\end{align}
\begin{align}
v_{i,t}^{\rm KG}
<&\frac{\sigma_i^{3}}{\left(\hat{\mu}_{b,t}-\hat{\mu}_{i,t}\right)^{2}}N_{i,t}^{-\frac{3}{2}}\left(N_{i,t}+1\right)^{-\frac{3}{2}}\phi\left(\frac{\hat{\mu}_{i,t}-\hat{\mu}_{b,t}}{\sigma_i}\sqrt{N_{i,t}\left(N_{i,t}+1\right)}\right)\nonumber\\
<&\frac{1}{\sqrt{2\pi}}\exp\left\{-\frac{\left(\mu_b-\mu_i-t^{-\frac{1}{4}}\right)^{2}}{2\sigma_i^{2}}N_{i,t}^{2}+\ln\left(\frac{\sigma_i^{3}N_{i,t}^{-3}}{\left(\mu_b-\mu_i-t^{-\frac{1}{4}}\right)^{2}}\right)\right\}\nonumber\\
<&\frac{1}{\sqrt{2\pi}}\exp\left\{-\frac{\left(\mu_b-\mu_i-t^{-\frac{1}{4}}\right)^{2}}{2\sigma_i^{2}}N_{i,t}^{2}\right\}	\triangleq\overline{v}_{i,t}^{\rm KG}.\label{value-kg-nb-up}
\end{align}
Similarly, for $\forall i\ne b$, $\forall t>T_3$, with a probability of at least $\left[1-q(t)\right]^{k}$,
\begin{align}
v_{b,t}^{\rm KG}>&\frac{1}{\sqrt{2\pi}}\exp\left\{-\left[\frac{\left(\mu_b-\max_{j\ne b}\mu_j+t^{-\frac{1}{4}}\right)^{2}}{2\sigma_b^{2}}+\frac{4k}{\sqrt{t}}+\frac{k\max\left\{\ln\left(\frac{27\delta_{\rm max}^{3}}{8\sigma_{\rm min}^{4}}\right),0\right\}}{t}\right]\left(N_{b,t}+1\right)^{2}\right\}	\triangleq\underline{v}_{b,t}^{\rm KG},\label{value-kg-b-low}\\
v_{b,t}^{\rm KG}<&\frac{1}{\sqrt{2\pi}}\exp\left\{-\frac{\left(\mu_b-\max_{j\ne b}\mu_j-t^{-\frac{1}{4}}\right)^{2}}{2\sigma_b^{2}}N_{b,t}^{2}\right\}\triangleq\overline{v}_{b,t}^{\rm KG}.\label{value-kg-b-up}
\end{align}

For $\forall i$, denote  $t_i=\min\left\{s\geq t\left|I_s=i\right.\right\}$ and $\tilde{t}_i=\max\left\{s\leq t\left|I_s=i\right.\right\}$. For $\forall i_1,i_2$, $i_1\ne i_2$, there exist two cases shown as follows:
\begin{itemize}
	\item When $t\leq t_{i_1}<t_{i_2}$, 
	\begin{align}\label{prop2-case1}
	\frac{N_{i_1,\tilde{t}_{i_2}}}{N_{i_2,\tilde{t}_{i_2}}+1}\leq\frac{N_{i_1,t}}{N_{i_2,t}}\leq\frac{N_{i_1,t_{i_1}}}{N_{i_2,t_{i_1}}}.
	\end{align}
	
	\item When $t\leq t_{i_2}<t_{i_1}$,
	\begin{align}\label{prop2-case2}
	\frac{N_{i_1,t_{i_2}}}{N_{i_2,t_{i_2}}}\leq\frac{N_{i_1,t}}{N_{i_2,t}}\leq\frac{N_{i_1,\tilde{t}_{i_1}}+1}{N_{i_2,\tilde{t}_{i_1}}}.
	\end{align}
\end{itemize}

We first focus on (\ref{prop2-case1}). In round $t_{i}$, $i\ne b$, $v_{i,t_{i}}^{\rm KG}>v_{b,t_{i}}^{\rm KG}$, and thus $\overline{v}_{i,t_{i}}^{\rm KG}>\underline{v}_{b,t_{i}}^{\rm KG}$. Based on (\ref{value-kg-nb-up}) and (\ref{value-kg-b-low}), for $\forall i\ne b$, $\forall t_i>T_3$, with a probability of at least $\left[1-q\left(t_i\right)\right]^{k}$,
\begin{align*}
\frac{\left(\mu_b-\mu_i-t_i^{-\frac{1}{4}}\right)^{2}}{2\sigma_i^{2}}N_{i,t_i}^{2}<\left[\frac{\left(\mu_b-\max_{j\ne b}\mu_j+t_i^{-\frac{1}{4}}\right)^{2}}{2\sigma_b^{2}}+\frac{4k}{\sqrt{t_i}}+\frac{k\max\left\{\ln\left(\frac{27\delta_{\rm max}^{3}}{8\sigma_{\rm min}^{4}}\right),0\right\}}{t_i}\right]\left(N_{b,t_i}+1\right)^{2},&
\end{align*}
that is,
\begin{align}
\frac{N_{i,t_i}}{N_{b,t_i}}<&\frac{\left(1+\frac{1}{N_{b,t_i}}\right)\sigma_i}{\mu_b-\mu_i-t_i^{-\frac{1}{4}}}\left(\frac{\left(\mu_b-\max_{j\ne b}\mu_j+t_i^{-\frac{1}{4}}\right)^{2}}{\sigma_b^{2}}+\frac{8k}{\sqrt{t_i}}+\frac{2k\max\left\{\ln\left(\frac{27\delta_{\rm max}^{3}}{8\sigma_{\rm min}^{4}}\right),0\right\}}{t_i}\right)^{\frac{1}{2}}\nonumber\\
\leq&\frac{\left(1+\left(\frac{t_i}{k}\right)^{-\frac{3}{4}}\right)\sigma_i}{\mu_b-\mu_i-t_i^{-\frac{1}{4}}}\left(\frac{\left(\mu_b-\max_{j\ne b}\mu_j+t_i^{-\frac{1}{4}}\right)^{2}}{\sigma_b^{2}}+\frac{8k}{\sqrt{t_i}}+\frac{2k\max\left\{\ln\left(\frac{27\delta_{\rm max}^{3}}{8\sigma_{\rm min}^{4}}\right),0\right\}}{t_i}\right)^{\frac{1}{2}}\nonumber\\
<&\frac{\left(1+\left(\frac{t}{k}\right)^{-\frac{3}{4}}\right)\sigma_i}{\mu_b-\mu_i-t^{-\frac{1}{4}}}\left(\frac{\left(\mu_b-\max_{j\ne b}\mu_j+t^{-\frac{1}{4}}\right)^{2}}{\sigma_b^{2}}+\frac{8k}{\sqrt{t}}+\frac{2k\max\left\{\ln\left(\frac{27\delta_{\rm max}^{3}}{8\sigma_{\rm min}^{4}}\right),0\right\}}{t}\right)^{\frac{1}{2}}\triangleq\overline{\rho}_{i,b,t}^{(1)},\label{prop2-sample-ratio-case1-ub-b-up}
\end{align}
The second inequality holds because $N_{b,t_i}\geq\left(\frac{t_i}{k}\right)^{\frac{3}{4}}$ for $\forall i$. The last inequality holds because $t_i\geq t$ for $\forall i$. Based on (\ref{prop2-case1}), for $\forall i\ne b$, $\forall t>T_3$, with a probability of at least $\left[1-q(t)\right]^{k}$, $\frac{N_{i,t}}{N_{b,t}}<\overline{\rho}_{i,b,t}$.

In round $\tilde{t}_b$, $v_{b,\tilde{t}_b}^{\rm KG}>v_{i,\tilde{t}_b}^{\rm KG}$, and thus $\overline{v}_{b,\tilde{t}_b}^{\rm KG}>\underline{v}_{i,\tilde{t}_b}^{\rm KG}$. Based on (\ref{value-kg-nb-low}) and (\ref{value-kg-b-up}),  for $\forall i\ne b$, $\forall \tilde{t}_b>T_3$, with a probability of at least $\left[1-q\left(\tilde{t}_b\right)\right]^{k}$,
\begin{align*}
\frac{\left(\mu_b-\max_{j\ne b}\mu_j-\tilde{t}_b^{-\frac{1}{4}}\right)^{2}}{2\sigma_b^{2}}N_{b,\tilde{t}_b}^{2}<\left[\frac{\left(\mu_b-\mu_i+\tilde{t}_b^{-\frac{1}{4}}\right)^{2}}{2\sigma_i^{2}}+\frac{4k}{\sqrt{\tilde{t}_b}}+\frac{k\max\left\{\ln\left(\frac{27\delta_{\rm max}^{3}}{8\sigma_{\rm min}^{4}}\right),0\right\}}{\tilde{t}_b}\right]\left(N_{i,\tilde{t}_b}+1\right)^{2},&
\end{align*}
that is,
\begin{align}\label{prop2-sample-ratio-case1-ub-b-low-1}
\frac{N_{i,\tilde{t}_b}+1}{N_{b,\tilde{t}_b}}>&\frac{\mu_b-\max_{j\ne b}\mu_j-\tilde{t}_b^{-\frac{1}{4}}}{\sigma_b}\left(\frac{\left(\mu_b-\mu_i+\tilde{t}_b^{-\frac{1}{4}}\right)^{2}}{\sigma_i^{2}}+\frac{8k}{\sqrt{\tilde{t}_b}}+\frac{2k\max\left\{\ln\left(\frac{27\delta_{\rm max}^{3}}{8\sigma_{\rm min}^{4}}\right),0\right\}}{\tilde{t}_b}\right)^{-\frac{1}{2}}.
\end{align}
Notice that for $\forall i\ne b$, 
\begin{align*}
\eta_{\tilde{t}_b}=
&\frac{8+8\sqrt{2k}\sigma_{\rm max}}{\delta_{\rm min}}\tilde{t}_b^{\frac{3}{4}}+\left(\frac{16+32\sqrt{2}\sigma_{\rm max}+32k\sigma_{\rm max}^{2}}{\delta_{\rm min}^{2}}+\frac{4\sigma_{\rm max}\sqrt{2k\max\left\{\ln\left(\frac{27\delta_{\rm max}^{3}}{8\sigma_{\rm min}^{4}}\right),0\right\}}}{\delta_{\rm min}}\right)\tilde{t}_b^{\frac{1}{2}}\\
&+\frac{\left(16\sigma_{\rm max}\sqrt{2k}+32k\sigma_{\rm max}^{2}\right)\sqrt{\max\left\{\ln\left(\frac{27\delta_{\rm max}^{3}}{8\sigma_{\rm min}^{4}}\right),0\right\}}}{\delta_{\rm min}^{2}}\tilde{t}_b^{\frac{1}{4}}+\frac{8k\sigma_{\rm max}^{2}\max\left\{\ln\left(\frac{27\delta_{\rm max}^{3}}{8\sigma_{\rm min}^{4}}\right),0\right\}}{\delta_{\rm min}^{2}}+2\\
&+\left(\frac{8}{\delta_{\rm min}}+\frac{8\sqrt{2k}\sigma_{\rm max}}{\delta_{\rm min}}\right)\tilde{t}_b^{-\frac{1}{4}}+\left(\frac{16+32\sqrt{2}\sigma_{\rm max}+32k\sigma_{\rm max}^{2}}{\delta_{\rm min}^{2}}+\frac{4\sigma_{\rm max}\sqrt{2k\max\left\{\ln\left(\frac{27\delta_{\rm max}^{3}}{8\sigma_{\rm min}^{4}}\right),0\right\}}}{\delta_{\rm min}}\right)\tilde{t}_b^{-\frac{1}{2}}\\
&+\frac{\left(16\sigma_{\rm max}\sqrt{2k}+32k\sigma_{\rm max}^{2}\right)\sqrt{\max\left\{\ln\left(\frac{27\delta_{\rm max}^{3}}{8\sigma_{\rm min}^{4}}\right),0\right\}}}{\delta_{\rm min}^{2}}\tilde{t}_b^{-\frac{3}{4}}+\frac{8k\sigma_{\rm max}^{2}\max\left\{\ln\left(\frac{27\delta_{\rm max}^{3}}{8\sigma_{\rm min}^{4}}\right),0\right\}}{\delta_{\rm min}^{2}}\tilde{t}_b^{-1},
\end{align*}
it holds that
\begin{align*}
&\frac{\left(\mu_b-\mu_i-\tilde{t}_b^{-\frac{1}{4}}\right)^{2}}{2\sigma_i^{2}}\left(N_{i,\tilde{t}_b}+\eta_{\tilde{t}_b}\right)^{2}\\
>&\left[\frac{\left(\mu_b-\max_{j\ne b}\mu_j+\tilde{t}_b^{-\frac{1}{4}}\right)^{2}}{2\sigma_b^{2}}+\frac{4k}{\sqrt{\tilde{t}_b}}+\frac{k\max\left\{\ln\left(\frac{27\delta_{\rm max}^{3}}{8\sigma_{\rm min}^{4}}\right),0\right\}}{\tilde{t}_b}\right]\left(N_{b,\tilde{t}_b}+1\right)^{2}.
\end{align*}
Then, for $t'=\min\left\{s\geq\tilde{t}_b\left|N_{i,s}\geq N_{i,\tilde{t}_b}+\eta_{\tilde{t}_b}, \forall i\ne b\right.\right\}$, $v_{b,\tilde{t}_b+1}^{\rm KG}>v_{i,t'}^{\rm KG}$ for $\forall i\ne b$. It indicates that 
\begin{align}\label{prop2-tilde-t-b-low}
\tilde{t}_b+(k-1)\eta_{t}\geq t,
\end{align}
where $\eta_t>\eta_{\tilde{t}_b}$,
\begin{align*}
\eta_{t}=&\frac{8+8\sqrt{2k}\sigma_{\rm max}}{\delta_{\rm min}}t^{\frac{3}{4}}+\left(\frac{16+32\sqrt{2}\sigma_{\rm max}+32k\sigma_{\rm max}^{2}}{\delta_{\rm min}^{2}}+\frac{4\sigma_{\rm max}\sqrt{2k\max\left\{\ln\left(\frac{27\delta_{\rm max}^{3}}{8\sigma_{\rm min}^{4}}\right),0\right\}}}{\delta_{\rm min}}\right)t^{\frac{1}{2}}\\
&+\frac{\left(16\sigma_{\rm max}\sqrt{2k}+32k\sigma_{\rm max}^{2}\right)\sqrt{\max\left\{\ln\left(\frac{27\delta_{\rm max}^{3}}{8\sigma_{\rm min}^{4}}\right),0\right\}}}{\delta_{\rm min}^{2}}t^{\frac{1}{4}}+\frac{8k\sigma_{\rm max}^{2}\max\left\{\ln\left(\frac{27\delta_{\rm max}^{3}}{8\sigma_{\rm min}^{4}}\right),0\right\}}{\delta_{\rm min}^{2}}+2\\
&+\frac{8+8\sqrt{2k}\sigma_{\rm max}}{\delta_{\rm min}}+\frac{16+32\sqrt{2}\sigma_{\rm max}+32k\sigma_{\rm max}^{2}}{\delta_{\rm min}^{2}}+\frac{4\sigma_{\rm max}\sqrt{2k\max\left\{\ln\left(\frac{27\delta_{\rm max}^{3}}{8\sigma_{\rm min}^{4}}\right),0\right\}}}{\delta_{\rm min}}\\
&+\frac{\left(16\sigma_{\rm max}\sqrt{2k}+32k\sigma_{\rm max}^{2}\right)\sqrt{\max\left\{\ln\left(\frac{27\delta_{\rm max}^{3}}{8\sigma_{\rm min}^{4}}\right),0\right\}}}{\delta_{\rm min}^{2}}+\frac{8k\sigma_{\rm max}^{2}\max\left\{\ln\left(\frac{27\delta_{\rm max}^{3}}{8\sigma_{\rm min}^{4}}\right),0\right\}}{\delta_{\rm min}^{2}}.
\end{align*}
Notice that $\exists T_4>T_3$, $\forall t>T_4$, $t-\left(k-1\right)\eta_t\geq\frac{3}{4}t$. Based on (\ref{prop2-sample-ratio-case1-ub-b-low-1}) and (\ref{prop2-tilde-t-b-low}), 
for $\forall i\ne b$, $\forall t>T_4$, with a probability of at least $\left[1-q\left(\frac{3}{4}t\right)\right]^{k}$,
\begin{align}
&\frac{N_{i,\tilde{t}_b}}{N_{b,\tilde{t}_b}+1}\nonumber\\
>&\frac{\mu_b-\max_{j\ne b}\mu_j-\left(\frac{3}{4}t\right)^{-\frac{1}{4}}}{\left(1+\left(\frac{3t}{4k}\right)^{-\frac{3}{4}}\right)^{2}\sigma_b}\left(\frac{\left(\mu_b-\mu_i+\left(\frac{3}{4}t\right)^{-\frac{1}{4}}\right)^{2}}{\sigma_i^{2}}+\frac{16k}{\sqrt{3t}}+\frac{8k\max\left\{\ln\left(\frac{27\delta_{\rm max}^{3}}{8\sigma_{\rm min}^{4}}\right),0\right\}}{3t}\right)^{-\frac{1}{2}}\triangleq\underline{\rho}_{i,b,t}^{(1)},\label{prop2-sample-ratio-case1-ub-b-low}
\end{align}
In view of (\ref{prop2-case1}), (\ref{prop2-sample-ratio-case1-ub-b-up}), and (\ref{prop2-sample-ratio-case1-ub-b-low}), for $\forall i\ne b$, $\forall \tilde{t}_b>T_4$, with a probability of at least $\left[1-q\left(\frac{3}{4}t\right)\right]^{k}$,
\begin{align*}
\underline{\rho}_{i,b,t}^{(1)}<\frac{N_{i,t}}{N_{b,t}}<\overline{\rho}_{i,b,t}^{(1)}.
\end{align*}

Next, we consider the case of (\ref{prop2-case2}). Following similar discussion as in the case of (\ref{prop2-case1}), we can derive that for $\forall i\ne b$, $\forall t>T_4$, with a probability of at least $\left[1-q\left(\frac{3}{4}t\right)\right]^{k}$,
\begin{align*}
\underline{\rho}_{i,b,t}^{(2)}<\frac{N_{i,t}}{N_{b,t}}<\overline{\rho}_{i,b,t}^{(2)},
\end{align*}
where 
\begin{align*}
\underline{\rho}_{i,b,t}^{(2)}=&\frac{\mu_b-\max_{j\ne b}\mu_j-t^{-\frac{1}{4}}}{\left(1+\left(\frac{t}{k}\right)^{-\frac{3}{4}}\right)\sigma_b}\left(\frac{\left(\mu_b-\mu_i+t^{-\frac{1}{4}}\right)^{2}}{\sigma_i^{2}}+\frac{8k}{\sqrt{t}}+\frac{2k\max\left\{\ln\left(\frac{27\delta_{\rm max}^{3}}{8\sigma_{\rm min}^{4}}\right),0\right\}}{t}\right)^{-\frac{1}{2}},\\
\overline{\rho}_{i,b,t}^{(2)}=&\frac{\left(1+\left(\frac{3t}{4k}\right)^{-\frac{3}{4}}\right)^{2}\sigma_i}{\mu_b-\mu_i-\left(\frac{3}{4}t\right)^{-\frac{1}{4}}}\left(\frac{\left(\mu_b-\max_{j\ne b}\mu_j+\left(\frac{3}{4}t\right)^{-\frac{1}{4}}\right)^{2}}{\sigma_b^{2}}+\frac{16k}{\sqrt{3t}}+\frac{8k\max\left\{\ln\left(\frac{27\delta_{\rm max}^{3}}{8\sigma_{\rm min}^{4}}\right),0\right\}}{3t}\right)^{\frac{1}{2}}.
\end{align*}
Denote by $\underline{\rho}_{i,b,t}=\min\left\{\underline{\rho}_{i,b,t}^{(1)},\underline{\rho}_{i,b,t}^{(2)}\right\}$, $\overline{\rho}_{i,b,t}=\max\left\{\overline{\rho}_{i,b,t}^{(1)},\overline{\rho}_{i,b,t}^{(2)}\right\}$. 
For $\forall i\ne b$, $\forall t>T_4$, with a probability of at least $\left[1-q\left(\frac{3}{4}t\right)\right]^{k}$, 
\begin{align*}
\underline{\rho}_{i,b,t}<\frac{N_{i,t}}{N_{b,t}}<\overline{\rho}_{i,b,t}.
\end{align*}
\hfill$\blacksquare$

\section{Proof of Theorem \ref{thm1}}

Based on Proposition \ref{prop2}, for $\forall t>T$, with a probability of at least $\left[1-q\left(\frac{3}{4}t\right)\right]^{k}$, 
\begin{align*}
\left(1+\sum_{j\ne b}\underline{\rho}_{i,b,t}\right)N_{b,t}\leq\sum_{i=1}^{k}N_{i,t}\leq\left(1+\sum_{j\ne b}\overline{\rho}_{i,b,t}\right)N_{b,t}.
\end{align*}
Notice that $\alpha_{i,t}=\frac{N_{b,t}}{\sum_{j=1}^{k}N_{j,t}}\frac{N_{i,t}}{N_{b,t}}$ for $\forall i\in\left\{1,\dots,k\right\}$. Then, 
\begin{align*}
\frac{1}{1+\sum_{i\ne b}\overline{\rho}_{i,b,t}}\leq \alpha_{b,t}\leq\frac{1}{1+\sum_{i\ne b}\underline{\rho}_{i,b,t}},~
\frac{\underline{\rho}_{i,b,t}}{1+\sum_{j\ne b}\overline{\rho}_{j,b,t}}\leq \alpha_{i,t}\leq\frac{\overline{\rho}_{i,b,t}}{1+\sum_{j\ne b}\underline{\rho}_{j,b,t}},~\forall i\ne b.
\end{align*}
\hfill$\blacksquare$

\section{Proof of Corollary \ref{cor1}}

For $\forall i$, denote by $L_{i,s}=\mathbbm{1}\left\{I_s=i\right\}$, $M_{i,s}=\mathbb{E}\left[L_{i,s}\left|L_{i,1},\dots,L_{i,s-1}\right.\right]-L_{i,s}$, $K_{i,t}=\sum_{s=1}^{t}M_{i,s}=\mathbb{E}\left[N_{i,t}\right]-N_{i,t}$, $V_t=\sum_{s=1}^{t}{\rm Var}\left[L_s\left|L_1,\dots,L_{s-1}\right.\right]$, $\tilde{M}_{i,s}=-M_{i,s}$, $\tilde{K}_{i,t}=\sum_{s=1}^{t}\tilde{M}_{i,s}=N_{i,t}-\mathbb{E}\left[N_{i,t}\right]$. Notice that $V_t\leq\sum_{s=1}^{t}\mathbb{P}\left(L_s\left|L_1,\dots,L_{s-1}\right.\right)=\mathbb{E}\left[N_{i,t}\right]$. According to Lemma \ref{lemma4}, $\mathbb{P}\left(N_{i,t}\leq\left(1-t^{-\frac{1}{8}}\right)\mathbb{E}\left[N_{i,t}\right]\right)\leq\exp\left\{-\frac{\mathbb{E}\left[N_{i,t}\right]}{2t^{\frac{1}{4}}+\frac{2}{3}t^{\frac{1}{8}}}\right\}$, $\mathbb{P}\left(N_{i,t}\geq\left(1+t^{-\frac{1}{8}}\right)\mathbb{E}\left[N_{i,t}\right]\right)\leq\exp\left\{-\frac{\mathbb{E}\left[N_{i,t}\right]}{2t^{\frac{1}{4}}+\frac{2}{3}t^{\frac{1}{8}}}\right\}$. Then, it holds almost surely that $\forall i$,
\begin{align}
&\left(1-\exp\left\{-\frac{1}{3\sqrt{k}}t^{\frac{1}{4}}\right\}\right)\left(1-t^{-\frac{1}{8}}\right)\mathbb{E}\left[N_{i,t}\right]+\exp\left\{-\frac{1}{2}t^{\frac{3}{4}}\right\}\label{thm1-number-low}\\
<&\left[1-\exp\left\{-\frac{\mathbb{E}\left[N_{i,t}\right]}{2t^{\frac{1}{4}}+\frac{2}{3}t^{\frac{1}{8}}}\right\}\right]\left(1-t^{-\frac{1}{8}}\right)\mathbb{E}\left[N_{i,t}\right]+\exp\left\{-\frac{\mathbb{E}\left[N_{i,t}\right]}{2t^{\frac{1}{4}}+\frac{2}{3}t^{\frac{1}{8}}}\right\}\nonumber\\
\leq& N_{i,t}\leq\left[1-\exp\left\{-\frac{\mathbb{E}\left[N_{i,t}\right]}{2t^{\frac{1}{4}}+\frac{2}{3}t^{\frac{1}{8}}}\right\}\right]\left(1+t^{-\frac{1}{8}}\right)\mathbb{E}\left[N_{i,t}\right]+t\exp\left\{-\frac{\mathbb{E}\left[N_{i,t}\right]}{2t^{\frac{1}{4}}+\frac{2}{3}t^{\frac{1}{8}}}\right\}\nonumber\\
<&\left(1-\exp\left\{-\frac{1}{2}t^{\frac{3}{4}}\right\}\right)\left(1+t^{-\frac{1}{8}}\right)\mathbb{E}\left[N_{i,t}\right]+t\exp\left\{-\frac{1}{3\sqrt{k}}t^{\frac{1}{4}}\right\}.\label{thm1-number-up}
\end{align}
According to Proposition \ref{prop2}, for arm $b$,
\begin{align}
&\frac{\left[1-q\left(\frac{3}{4}t\right)\right]^{k}}{1+\sum_{i\ne b}\overline{\rho}_{i,b,t}}t\label{thm1-expected-number-b-low}\\
<&\left[1-q\left(\frac{3}{4}t\right)\right]^{k}\mathbb{E}\left[N_{b,t}\left|N_{b,t}\geq\frac{t}{1+\sum_{j\ne b}\overline{\rho}_{j,b,t}}\right.\right]+\left[q\left(\frac{3}{4}t\right)\right]^{k}\mathbb{E}\left[N_{b,t}\left|N_{b,t}<\frac{t}{1+\sum_{j\ne b}\overline{\rho}_{j,b,t}}\right.\right]\nonumber\\
\leq&\mathbb{E}\left[N_{b,t}\right]\nonumber\\
\leq&\mathbb{E}\left[N_{b,t}\left|N_{b,t}\leq\frac{t}{1+\sum_{j\ne b}\underline{\rho}_{j,b,t}}\right.\right]+kq\left(\frac{3}{4}t\right)\mathbb{E}\left[N_{b,t}\left|N_{b,t}>\frac{t}{1+\sum_{j\ne b}\underline{\rho}_{j,b,t}}\right.\right]\nonumber\\
<&\frac{t}{1+\sum_{j\ne b}\underline{\rho}_{j,b,t}}+kq\left(\frac{3}{4}t\right)t.\label{thm1-expected-number-b-up}
\end{align}
Meanwhile, for $\forall i\ne b$,
\begin{align}
&\frac{\left[1-q\left(\frac{3}{4}t\right)\right]^{k}\underline{\rho}_{i,b,t}}{1+\sum_{j\ne b}\overline{\rho}_{j,b,t}}t\label{thm1-expected-number-nb-low}\\
<&\left[1-q\left(\frac{3}{4}t\right)\right]^{k}\mathbb{E}\left[N_{i,t}\left|N_{i,t}\geq\frac{\underline{\rho}_{i,b,t}t}{1+\sum_{j\ne b}\overline{\rho}_{j,b,t}}\right.\right]+\left[q\left(\frac{3}{4}t\right)\right]^{k}\mathbb{E}\left[N_{i,t}\left|N_{i,t}<\frac{\underline{\rho}_{i,b,t}t}{1+\sum_{j\ne b}\overline{\rho}_{j,b,t}}\right.\right]\nonumber\\
\leq&\mathbb{E}\left[N_{i,t}\right]\leq\mathbb{E}\left[N_{i,t}\left|N_{i,t}\leq\frac{\overline{\rho}_{i,b,t}t}{1+\sum_{j\ne b}\underline{\rho}_{j,b,t}}\right.\right]+kq\left(\frac{3}{4}t\right)\mathbb{E}\left[N_{i,t}\left|N_{i,t}>\frac{\overline{\rho}_{i,b,t}t}{1+\sum_{j\ne b}\underline{\rho}_{j,b,t}}\right.\right]\nonumber\\
<&\frac{\overline{\rho}_{i,b,t}t}{1+\sum_{j\ne b}\underline{\rho}_{j,b,t}}+kq\left(\frac{3}{4}t\right)t.\label{thm1-expected-number-nb-up}
\end{align}
Thus, for $\forall i\ne b$, $\mathbb{P}\left(\lim_{t\to\infty}\frac{N_{i,t}}{N_{b,t}}=\frac{\sigma_i}{\sigma_b}\frac{\mu_b-\max_{j\ne b}\mu_j}{\mu_b-\mu_i}\right)=1$ can be obtained by combining (\ref{thm1-number-low}), (\ref{thm1-number-up}) with (\ref{thm1-expected-number-b-low}), (\ref{thm1-expected-number-b-up}). Similarly,  for $\forall i_1,i_2\ne b$, $\mathbb{P}\left(\lim_{t\to\infty}\frac{N_{i_1,t}}{N_{i_2,t}}=\frac{\sigma_{i_1}}{\sigma_{i_2}}\frac{\mu_b-\mu_{i_2}}{\mu_b-\mu_{i_1}}\right)=1$ can be obtained by combining (\ref{thm1-number-low}), (\ref{thm1-number-up}) with (\ref{thm1-expected-number-nb-low}), (\ref{thm1-expected-number-nb-up}).
\hfill$\blacksquare$

\section{Proof of Theorem \ref{thm2}}\label{thm2-proof}

Suppose that all the random variables are defined on a probability space $\left(\Omega,\mathcal{F},\mathbb{P}\right)$. Denote by $\omega$ any sample path in $\Omega$. According to the Strong Law of Large Numbers, there exists a measurable set $\Omega_1\subseteq\Omega$ such that $\mathbb{P}\left(\omega\in\Omega_1\right)=1$, $\hat{\mu}_{i,t}\to\mu_i$ as $t\to\infty$, $\hat{\mu}_{i_1,t_1}\ne\hat{\mu}_{i_2,t_2}$ for all $i_1\ne i_2$ and $t_1,t_2\geq 1$, and $N_{i,t}\geq\left(\frac{t}{k}\right)^{\frac{3}{4}}$ for $\forall \omega\in\Omega_1$, $\forall i$.  

Note that $J_t\triangleq J_t\left(\omega\right)$, $\hat{\mu}_{i,t}\triangleq\hat{\mu}_{i,t}\left(\omega\right)$, $N_{i,t}\triangleq N_{i,t}\left(\omega\right)$ for $\forall \omega\in\Omega$, $\forall i$, $\forall t$. We use $J_t$, $\hat{\mu}_{i,t}$ and $N_{i,t}$ in the proof below. 
\begin{align}
e_t=&\mathbb{P}\left(\left.J_t\ne b\right|\omega\in\Omega_1\right)\mathbb{P}\left(\omega\in\Omega_1\right)+\mathbb{P}\left(\left.J_t\ne b\right|\omega\in\Omega\setminus\Omega_1\right)\mathbb{P}\left(\omega\in\Omega\setminus\Omega_1\right)\nonumber\\
=&\mathbb{P}\left(\left.J_t\ne b\right|\omega\in\Omega_1\right).\label{PE_new}
\end{align}
According to Theorem \ref{thm1}, $\exists T>T_0$, for $\forall t>T$, there exists a measurable set $\Omega_{2,t}\subseteq\Omega_1$ such that $\mathbb{P}\left(\omega\in\Omega_{2,t}\right)\geq\left[1-q\left(\frac{3}{4}t\right)\right]^{k}$, and for $\forall \omega\in\Omega_{2,t}$, 
\begin{equation}\label{thm2_sample_bounds}
\begin{aligned}
&\frac{t}{1+\sum_{i\ne b}\overline{\rho}_{i,b,t}}\leq N_{b,t}\leq\frac{t}{1+\sum_{i\ne b}\underline{\rho}_{i,b,t}},\\
&\frac{\underline{\rho}_{i,b,t}t}{1+\sum_{j\ne b}\overline{\rho}_{j,b,t}}\leq N_{i,t}\leq\frac{\overline{\rho}_{i,b,t}t}{1+\sum_{j\ne b}\underline{\rho}_{j,b,t}},\forall i\ne b.
\end{aligned}
\end{equation}

Inspired by \cite{wu2018}, under the KG algorithm, for $\forall\omega\in\Omega_1$, $\forall t$, if the event
\begin{align*}
\mathcal{E}_t=\left\{\hat{\mu}_{b,t}>\mu_b-\frac{\delta_{\rm min}}{2}\right\}\bigcap\left\{\bigcap_{i\ne b}\left\{\hat{\mu}_{i,t}\leq\mu_b-\frac{\delta_{\rm min}}{2}\right\}\right\}
\end{align*}
occurs in round $t$, then the correct selection occurs in round $t$ regardless of the exact values of $N_{i,t}$'s. For $\forall t$, denote by $\mathcal{E}_t^c$ the complementary event of $\mathcal{E}_t$. Based on (\ref{PE_new}),
\begin{align}
e_t\leq&\mathbb{P}\left(\left.\mathcal{E}_t^c\right|\omega\in\Omega_1\right)
\leq\mathbb{P}\left(\left.\hat{\mu}_{b,t}\leq\mu_b-\frac{\delta_{\rm min}}{2}\right|\omega\in\Omega_1\right)+\sum_{i\ne b}\mathbb{P}\left(\left.\hat{\mu}_{i,t}>\mu_b-\frac{\delta_{\rm min}}{2}\right|\omega\in\Omega_1\right)\nonumber\\
=&\mathbb{P}\left(\left.\hat{\mu}_{b,t}\leq\mu_b-\frac{\delta_{\rm min}}{2}\right|\omega\in\Omega_{2,t}\right)\mathbb{P}\left(\omega\in\Omega_{2,t}\right)\\
&+\mathbb{P}\left(\left.\hat{\mu}_{b,t}\leq\mu_b-\frac{\delta_{\rm min}}{2}\right|\omega\in\Omega_1\setminus\Omega_{2,t}\right)\mathbb{P}\left(\omega\in\Omega_1\setminus\Omega_{2,t}\right)\nonumber\\
&+\sum_{i\ne b}\left[\mathbb{P}\left(\left.\hat{\mu}_{i,t}>\mu_b-\frac{\delta_{\rm min}}{2}\right|\omega\in\Omega_{2,t}\right)\mathbb{P}\left(\omega\in\Omega_{2,t}\right)\right.\nonumber\\
&~~~~~~~~~\left.+\mathbb{P}\left(\left.\hat{\mu}_{i,t}>\mu_b-\frac{\delta_{\rm min}}{2}\right|\omega\in\Omega_1\setminus\Omega_{2,t}\right)\mathbb{P}\left(\omega\in\Omega_1\setminus\Omega_{2,t}\right)\right]\label{PE_upper-0}\\
\leq&\mathbb{P}\left(\left.\hat{\mu}_{b,t}\leq\mu_b-\frac{\delta_{\rm min}}{2}\right|\omega\in\Omega_{2,t}\right)+\mathbb{P}\left(\left.\hat{\mu}_{b,t}\leq\mu_b-\frac{\delta_{\rm min}}{2}\right|\omega\in\Omega_1\setminus\Omega_{2,t}\right)\mathbb{P}\left(\omega\in\Omega_1\setminus\Omega_{2,t}\right)\nonumber\\
&+\sum_{i\ne b}\left[\mathbb{P}\left(\left.\hat{\mu}_{i,t}>\mu_b-\frac{\delta_{\rm min}}{2}\right|\omega\in\Omega_{2,t}\right)\right.\nonumber\\
&~~~~~~~~~\left.+\mathbb{P}\left(\left.\hat{\mu}_{i,t}>\mu_b-\frac{\delta_{\rm min}}{2}\right|\omega\in\Omega_1\setminus\Omega_{2,t}\right)\mathbb{P}\left(\omega\in\Omega_1\setminus\Omega_{2,t}\right)\right],\label{PE_upper}
\end{align}
where the second inequality holds based on the Bonferroni inequality \cite{bonferroni1936}, (\ref{PE_upper-0}) holds based on the law of total probability, (\ref{PE_upper}) holds because $\mathbb{P}\left(\omega\in\Omega_{2,t}\right)\leq 1$ for $\forall t$.

For $\forall\omega\in\Omega_1$, $\forall t$, if given the normal rewards $X_{I_1,1},\dots,X_{I_t,t}$ and $N_{i,t}=n_i$ for $\forall i$, $\hat{\mu}_{i,t}$ follows a normal distribution with mean $\mu_i$ and variance $\frac{\sigma_i^2}{n_i}$ \cite{dekking2005}. Then, 
\begin{align}
\mathbb{P}\left(\left.\hat{\mu}_{b,t}\leq\mu_b-\frac{\delta_{\rm min}}{2}\right|\omega\in\Omega_1\right)=&\Phi\left(\frac{\delta_{\rm min}}{2\sigma_b}\sqrt{n_b}\right)\leq\frac{\sqrt{2}\sigma_b}{\sqrt{\pi n_b}\delta_{\rm min}}\exp\left\{-\frac{\delta_{\rm min}^2}{8\sigma_b^2}n_i\right\},\label{PE_upper-1}\\
\mathbb{P}\left(\left.\hat{\mu}_{i,t}>\mu_b-\frac{\delta_{\rm min}}{2}\right|\omega\in\Omega_1\right)=&\Phi\left(\frac{\mu_b-\mu_i-\frac{\delta_{\rm min}}{2}}{\sigma_i}\sqrt{n_i}\right)\nonumber\\
\leq&\frac{\sigma_i}{\sqrt{2\pi n_i}\left(\mu_b-\mu_i-\frac{\delta_{\rm min}}{2}\right)}\exp\left\{-\frac{\left(\mu_b-\mu_i-\frac{\delta_{\rm min}}{2}\right)^2}{2\sigma_i^2}n_i\right\},\label{PE_upper-2}
\end{align}
where $\Phi(\cdot)$ is the cumulative density function of standard normal distribution, the inequalities in (\ref{PE_upper-1}) and (\ref{PE_upper-2}) hold because according to \cite{gordon1941}, $\forall x>0$,
\begin{align}\label{gordon-ineq}
\frac{x}{\sqrt{2\pi}\left(1+x^{2}\right)}\exp\left\{-\frac{x^{2}}{2}\right\}\leq\Phi(x)\leq\frac{1}{\sqrt{2\pi}x}\exp\left\{-\frac{x^{2}}{2}\right\}.
\end{align}

By combining (\ref{thm2_sample_bounds}) with (\ref{PE_upper-1}), (\ref{PE_upper-2}), for $\forall t>T$,
\begin{align}
&\mathbb{P}\left(\left.\hat{\mu}_{b,t}\leq\mu_b-\frac{\delta_{\rm min}}{2}\right|\omega\in\Omega_{2,t}\right)\leq\frac{\sigma_b\sqrt{2\left(1+\sum_{i\ne b}\overline{\rho}_{i,b,t}\right)}}{\delta_{\rm min}\sqrt{\pi t}}\exp\left\{-\frac{\delta_{\rm min}^2}{8\sigma_b^2\left(1+\sum_{i\ne b}\overline{\rho}_{i,b,t}\right)}t\right\},\label{PE_upper-1-2}\\
&\mathbb{P}\left(\left.\hat{\mu}_{i,t}>\mu_b-\frac{\delta_{\rm min}}{2}\right|\omega\in\Omega_{2,t}\right)\leq\frac{\sigma_i\sqrt{1+\sum_{i\ne b}\overline{\rho}_{i,b,t}}}{\left(\mu_b-\mu_i-\frac{\delta_{\rm min}}{2}\right)\sqrt{2\pi\underline{\rho}_{i,b,t} t}}\exp\left\{-\frac{\left(\mu_b-\mu_i-\frac{\delta_{\rm min}}{2}\right)^2\underline{\rho}_{i,b,t}}{2\sigma_i^2\left(1+\sum_{i\ne b}\overline{\rho}_{i,b,t}\right)}t\right\}.\label{PE_upper-2-2}
\end{align}
For $\forall t$, $\forall \omega\in\Omega_1\setminus\Omega_{2,t}$, it holds that $N_{i,t}\geq\left(\frac{t}{k}\right)^{\frac{3}{4}}$, $\forall i$. By combining it with (\ref{PE_upper-1}), (\ref{PE_upper-2}),
\begin{align}
&\mathbb{P}\left(\left.\hat{\mu}_{b,t}\leq\mu_b-\frac{\delta_{\rm min}}{2}\right|\omega\in\Omega_1\setminus\Omega_{2,t}\right)\leq\frac{\sqrt{2} k^{\frac{3}{8}}\sigma_b}{\sqrt{\pi}\delta_{\rm min}t^{\frac{3}{8}}}\exp\left\{-\frac{\delta_{\rm min}^2}{8\sigma_b^2 k^{\frac{3}{4}}}t^{\frac{3}{4}}\right\},\label{PE_upper-1-3}\\
&\mathbb{P}\left(\left.\hat{\mu}_{i,t}>\mu_b-\frac{\delta_{\rm min}}{2}\right|\omega\in\Omega_1\setminus\Omega_{2,t}\right)\leq\frac{k^{\frac{3}{8}}\sigma_i}{\sqrt{2\pi}\left(\mu_b-\mu_i-\frac{\delta_{\rm min}}{2}\right)t^{\frac{3}{8}}}\exp\left\{-\frac{\left(\mu_b-\mu_i-\frac{\delta_{\rm min}}{2}\right)^2}{2\sigma_i^2k^{\frac{3}{4}}}t^{\frac{3}{4}}\right\}.\label{PE_upper-2-3}
\end{align}
In addition, for $\forall t>T$,
\begin{align}\label{thm2-prob-comple-upper}
\mathbb{P}\left(\omega\in\Omega_1\setminus\Omega_{2,t}\right)=1-\mathbb{P}\left(\omega\in\Omega_{2,t}\right)\leq 1-\left[1-q\left(\frac{3}{4}t\right)\right]^{k}.
\end{align}
By combining (\ref{PE_upper}), (\ref{PE_upper-1-2}), (\ref{PE_upper-1-3}), (\ref{PE_upper-2-2}),
(\ref{PE_upper-2-3}) and (\ref{thm2-prob-comple-upper}), we can obtain the upper bound of $e_t$.
\begin{align}
e_t\leq&\frac{\sigma_b\sqrt{2\left(1+\sum_{i\ne b}\overline{\rho}_{i,b,t}\right)}}{\delta_{\rm min}\sqrt{\pi t}}\exp\left\{-\frac{\delta_{\rm min}^2}{8\sigma_b^2\left(1+\sum_{i\ne b}\overline{\rho}_{i,b,t}\right)}t\right\}\nonumber\\
&+\frac{\sqrt{2} k^{\frac{3}{8}}\sigma_b}{\sqrt{\pi}\delta_{\rm min}t^{\frac{3}{8}}}\left[1-\left[1-q\left(\frac{3}{4}t\right)\right]^{k}\right]\exp\left\{-\frac{\delta_{\rm min}^2}{8\sigma_b^2 k^{\frac{3}{4}}}t^{\frac{3}{4}}\right\}\nonumber\\
&+\sum_{i\ne b}\left\{\rule{0em}{10mm}\frac{\sigma_i\sqrt{\left(1+\sum_{i\ne b}\overline{\rho}_{i,b,t}\right)}}{\left(\mu_b-\mu_i-\frac{\delta_{\rm min}}{2}\right)\sqrt{2\pi\underline{\rho}_{i,b,t} t}}\right.\exp\left\{-\frac{\left(\mu_b-\mu_i-\frac{\delta_{\rm min}}{2}\right)^2\underline{\rho}_{i,b,t}}{2\sigma_i^2\left(1+\sum_{i\ne b}\overline{\rho}_{i,b,t}\right)}t\right\}\nonumber\\
&~~~~~~~~~~~~~~\left.+\frac{k^{\frac{3}{8}}\sigma_i\left[1-\left[1-q\left(\frac{3}{4}t\right)\right]^{k}\right]}{\sqrt{2\pi}\left(\mu_b-\mu_i-\frac{\delta_{\rm min}}{2}\right)t^{\frac{3}{8}}}\exp\left\{-\frac{\left(\mu_b-\mu_i-\frac{\delta_{\rm min}}{2}\right)^2}{2\sigma_i^2k^{\frac{3}{4}}}t^{\frac{3}{4}}\right\}\rule{0em}{10mm}\right\}.\label{PE_upper_bound}
\end{align}

Furthermore, following similar discussion as used in (\ref{PE_new}),
\begin{align}\label{SR_new}
r_t=\mathbb{E}\left[\left.\mu_b-\mu_{J_t}\right|\omega\in\Omega_1\right].
\end{align}
Based on similar discussion in \cite{Audibert2010}, for $\forall t$,
\begin{align}
&\delta_{\rm min}\mathbb{P}\left(\left.J_t\ne b\right|\omega\in\Omega_1\right)\nonumber\\
\leq&\mathbb{E}\left[\left.\mu_b-\mu_{J_t}\right|\omega\in\Omega_1\right]=\sum_{i\ne b}\left(\mu_b-\mu_i\right)\mathbb{P}\left(\left.J_t=i\right|\omega\in\Omega_1\right)\label{SR_new_bound_0}\\
\leq&\delta_{\rm max}\mathbb{P}\left(\left.J_t\ne b\right|\omega\in\Omega_1\right).\nonumber
\end{align}
By combining (\ref{PE_new}), (\ref{PE_upper_bound}) and (\ref{SR_new_bound_0}), we can obtain the upper bound of $r_t$. 
\begin{equation*}
\begin{aligned}
r_t\leq&\frac{\delta_{\rm max}\sigma_b\sqrt{2\left(1+\sum_{i\ne b}\overline{\rho}_{i,b,t}\right)}}{\delta_{\rm min}\sqrt{\pi t}}\exp\left\{-\frac{\delta_{\rm min}^2}{8\sigma_b^2\left(1+\sum_{i\ne b}\overline{\rho}_{i,b,t}\right)}t\right\}\\
&+\frac{\sqrt{2} k^{\frac{3}{8}}\delta_{\rm max}\sigma_b}{\sqrt{\pi}\delta_{\rm min}t^{\frac{3}{8}}}\left[1-\left[1-q\left(\frac{3}{4}t\right)\right]^{k}\right]\exp\left\{-\frac{\delta_{\rm min}^2}{8\sigma_b^2 k^{\frac{3}{4}}}t^{\frac{3}{4}}\right\}\\
&+\sum_{i\ne b}\left\{\rule{0em}{10mm}\frac{\delta_{\rm max}\sigma_i\sqrt{1+\sum_{i\ne b}\overline{\rho}_{i,b,t}}}{\left(\mu_b-\mu_i-\frac{\delta_{\rm min}}{2}\right)\sqrt{2\pi\underline{\rho}_{i,b,t} t}}\right.\exp\left\{-\frac{\left(\mu_b-\mu_i-\frac{\delta_{\rm min}}{2}\right)^2\underline{\rho}_{i,b,t}}{2\sigma_i^2\left(1+\sum_{i\ne b}\overline{\rho}_{i,b,t}\right)}t\right\}\\
&~~~~~~~~~~~~~~+\frac{k^{\frac{3}{8}}\delta_{\rm max}\sigma_i\left[1-\left[1-q\left(\frac{3}{4}t\right)\right]^{k}\right]}{\sqrt{2\pi}\left(\mu_b-\mu_i-\frac{\delta_{\rm min}}{2}\right)t^{\frac{3}{8}}}\left.\exp\left\{-\frac{\left(\mu_b-\mu_i-\frac{\delta_{\rm min}}{2}\right)^2}{2\sigma_i^2k^{\frac{3}{4}}}t^{\frac{3}{4}}\right\}\rule{0em}{10mm}\right\}.
\end{aligned}
\end{equation*}
\hfill$\blacksquare$

\section{Proof of Proposition \ref{prop3}}\label{prop3-proof}

Similar to the discussion in Section \ref{thm2-proof}, there exists a measurable set $\Omega_1\subseteq\Omega$ such that $\mathbb{P}\left(\omega\in\Omega_1\right)=1$, $\hat{\mu}_{i,t}\to\mu_i$ as $t\to\infty$, $\hat{\mu}_{i_1,t_1}\ne\hat{\mu}_{i_2,t_2}$ for all $i_1\ne i_2$ and $t_1,t_2\geq 1$, and $N_{i,t}\geq\left(\frac{t}{k}\right)^{\frac{3}{4}}$ for $\forall \omega\in\Omega_1$, $\forall i$. In addition, $\exists T>T_0$, for $\forall t>T$, there exists a measurable set $\Omega_{2,t}\subseteq\Omega_1$ such that $\mathbb{P}\left(\omega\in\Omega_{2,t}\right)\geq\left[1-q\left(\frac{3}{4}t\right)\right]^{k}$, and for $\forall \omega\in\Omega_{2,t}$,  (\ref{thm2_sample_bounds}) holds. Notice that under the KG algorithm, for $\forall \omega\in\Omega_1$, $\forall t$, if there exists some $j\ne b$ and the event
\begin{align*}
\tilde{\mathcal{E}}_{j,t}=\left\{\hat{\mu}_{j,t}>\mu_j+\frac{\delta_{\rm min}}{2}\right\}\bigcap\left\{\hat{\mu}_{b,t}\leq\mu_j+\frac{\delta_{\rm min}}{2}\right\}
\end{align*}
occurs in round $t$, then the false selection occurs in round $t$ regardless of the exact values of $N_{i,t}$'s. Based on (\ref{PE_new}), $\exists j\ne b$, 
\begin{align}
e_t\geq&\mathbb{P}\left(\left.\tilde{\mathcal{E}}_{j,t}\right|\omega\in\Omega_1\right)
\geq\mathbb{P}\left(\left.\hat{\mu}_{j,t}>\mu_j+\frac{\delta_{\rm min}}{2}\right|\omega\in\Omega_1\right)\mathbb{P}\left(\left.\hat{\mu}_{b,t}\leq\mu_j+\frac{\delta_{\rm min}}{2}\right|\omega\in\Omega_1\right)\nonumber\\
=&\left[\mathbb{P}\left(\left.\hat{\mu}_{j,t}>\mu_j+\frac{\delta_{\rm min}}{2}\right|\omega\in\Omega_{2,t}\right)\mathbb{P}\left(\omega\in\Omega_{2,t}\right)\right.\nonumber\\
&~~\left.+\mathbb{P}\left(\left.\hat{\mu}_{j,t}>\mu_j+\frac{\delta_{\rm min}}{2}\right|\omega\in\Omega_1\setminus\Omega_{2,t}\right)\mathbb{P}\left(\omega\in\Omega_1\setminus\Omega_{2,t}\right)\right]\nonumber\\
&\cdot\left[\mathbb{P}\left(\left.\hat{\mu}_{b,t}\leq\mu_j+\frac{\delta_{\rm min}}{2}\right|\omega\in\Omega_{2,t}\right)\mathbb{P}\left(\omega\in\Omega_{2,t}\right)\right.\nonumber\\
&~~~~\left.+\mathbb{P}\left(\left.\hat{\mu}_{b,t}\leq\mu_j+\frac{\delta_{\rm min}}{2}\right|\omega\in\Omega_1\setminus\Omega_{2,t}\right)\mathbb{P}\left(\omega\in\Omega_1\setminus\Omega_{2,t}\right)\right]\label{PE_lower-1}\\
\geq& \mathbb{P}\left(\left.\hat{\mu}_{j,t}>\mu_j+\frac{\delta_{\rm min}}{2}\right|\omega\in\Omega_{2,t}\right)\mathbb{P}\left(\left.\hat{\mu}_{b,t}\leq\mu_j+\frac{\delta_{\rm min}}{2}\right|\omega\in\Omega_{2,t}\right)\left[\mathbb{P}\left(\omega\in\Omega_{2,t}\right)\right]^2,\label{PE_lower-2}
\end{align}
where (\ref{PE_lower-1}) holds based on the law of total probability, (\ref{PE_lower-2}) holds because $\mathbb{P}\left(\left.\hat{\mu}_{j,t}>\mu_j+\frac{\delta_{\rm min}}{2}\right|\right.$\\
$\left.\omega\in\Omega_1\setminus\Omega_{2,t}\right)\geq0$, $\mathbb{P}\left(\left.\hat{\mu}_{b,t}\leq\mu_j+\frac{\delta_{\rm min}}{2}\right|\omega\in\Omega_1\setminus\Omega_{2,t}\right)\geq 0$, $\mathbb{P}\left(\omega\in\Omega_1\setminus\Omega_{2,t}\right)\geq 0$. Similar to (\ref{PE_upper-1-2}) and (\ref{PE_upper-2-2}), we can derive that for $j\ne b$, $\forall t>T$, 
\begin{align}
&\mathbb{P}\left(\left.\hat{\mu}_{j,t}>\mu_j+\frac{\delta_{\rm min}}{2}\right|\omega\in\Omega_{2,t}\right)\nonumber\\
\geq&\frac{1}{\sqrt{2\pi}}\frac{\frac{\delta_{\rm min}}{2\sigma_j}\sqrt{\frac{\underline{\rho}_{j,b,t}t}{1+\sum_{i\ne b}\overline{\rho}_{i,b,t}}}}{1+\frac{\delta_{\rm min}^2\overline{\rho}_{j,b,t}t}{4\sigma_j^2\left(1+\sum_{i\ne b}\underline{\rho}_{i,b,t}\right)}}\exp\left\{-\frac{\delta_{\rm min}^2\overline{\rho}_{j,b,t}t}{8\sigma_j^2\left(1+\sum_{i\ne b}\underline{\rho}_{i,b,t}\right)}\right\},\label{PE_lower-1-2}\\
&\mathbb{P}\left(\left.\hat{\mu}_{b,t}\leq\mu_j+\frac{\delta_{\rm min}}{2}\right|\omega\in\Omega_{2,t}\right)\nonumber\\
\geq&\frac{1}{\sqrt{2\pi}}\frac{\frac{\left(\mu_b-\mu_i-\frac{\delta_{\rm min}}{2}\right)\sqrt{t}}{\sigma_b\sqrt{1+\sum_{i\ne b}\overline{\rho}_{i,b,t}}}}{1+\frac{\left(\mu_b-\mu_i-\frac{\delta_{\rm min}}{2}\right)^2 t}{\sigma_b^2\left(1+\sum_{i\ne b}\underline{\rho}_{i,b,t}\right)}}\exp\left\{-\frac{\left(\mu_b-\mu_i-\frac{\delta_{\rm min}}{2}\right)^2 t}{2\sigma_b^2\left(1+\sum_{i\ne b}\underline{\rho}_{i,b,t}\right)}\right\}.\label{PE_lower-2-2}
\end{align}
By combining (\ref{PE_lower-2}), (\ref{PE_lower-1-2}) and (\ref{PE_lower-2-2}), we can obtain the lower bound of $e_t$.
\begin{equation}\label{PE_lower_bound}
\begin{aligned}
&e_t\geq\frac{\left[1-q\left(\frac{3}{4}t\right)\right]^{2k}}{2\pi}\min_{j\ne b}\left\{\rule{0em}{10mm}\frac{\frac{\delta_{\rm min}}{2\sigma_j}\sqrt{\frac{\underline{\rho}_{j,b,t}t}{1+\sum_{i\ne b}\overline{\rho}_{i,b,t}}}}{1+\frac{\delta_{\rm min}^2\overline{\rho}_{j,b,t}t}{4\sigma_j^2\left(1+\sum_{i\ne b}\underline{\rho}_{i,b,t}\right)}}\frac{\frac{\left(\mu_b-\mu_j-\frac{\delta_{\rm min}}{2}\right)\sqrt{t}}{\sigma_b\sqrt{1+\sum_{i\ne b}\overline{\rho}_{i,b,t}}}}{1+\frac{\left(\mu_b-\mu_j-\frac{\delta_{\rm min}}{2}\right)^2 t}{\sigma_b^2\left(1+\sum_{i\ne b}\underline{\rho}_{i,b,t}\right)}}\right.\\
&~~~~~~~~~~~~~~~~~~~~~~~~~~~~~~~~~\cdot\left.\exp\left\{-\frac{\delta_{\rm min}^2\overline{\rho}_{j,b,t}t}{8\sigma_j^2\left(1+\sum_{i\ne b}\underline{\rho}_{i,b,t}\right)}\right\}\exp\left\{-\frac{\left(\mu_b-\mu_j-\frac{\delta_{\rm min}}{2}\right)^2 t}{2\sigma_b^2\left(1+\sum_{i\ne b}\underline{\rho}_{i,b,t}\right)}\right\}\rule{0em}{10mm}\right\}.
\end{aligned}
\end{equation}
Furthermore, based on (\ref{SR_new_bound_0}), $\mathbb{E}\left[\left.\mu_b-\mu_{J_t}\right|\omega\in\Omega_1\right]\geq\delta_{\rm min}\mathbb{P}\left(\left.J_t\ne b\right|\omega\in\Omega_1\right)$. By combining it with (\ref{PE_new}) and (\ref{PE_lower_bound}), we can obtain the lower bound of $r_t$.
\begin{align*}
&r_t\geq\frac{\left[1-q\left(\frac{3}{4}t\right)\right]^{2k}\delta_{\rm min}}{2\pi}\min_{j\ne b}\left\{\rule{0em}{10mm}\frac{\frac{\delta_{\rm min}}{2\sigma_j}\sqrt{\frac{\underline{\rho}_{j,b,t}t}{1+\sum_{i\ne b}\overline{\rho}_{i,b,t}}}}{1+\frac{\delta_{\rm min}^2\overline{\rho}_{j,b,t}t}{4\sigma_j^2\left(1+\sum_{i\ne b}\underline{\rho}_{i,b,t}\right)}}\frac{\frac{\left(\mu_b-\mu_j-\frac{\delta_{\rm min}}{2}\right)\sqrt{t}}{\sigma_b\sqrt{1+\sum_{i\ne b}\overline{\rho}_{i,b,t}}}}{1+\frac{\left(\mu_b-\mu_j-\frac{\delta_{\rm min}}{2}\right)^2 t}{\sigma_b^2\left(1+\sum_{i\ne b}\underline{\rho}_{i,b,t}\right)}}\right.\\
&~~~~~~~~~~~~~~~~~~~~~~~~~~~~~~~~~\cdot\left.\exp\left\{-\frac{\delta_{\rm min}^2\overline{\rho}_{j,b,t}t}{8\sigma_j^2\left(1+\sum_{i\ne b}\underline{\rho}_{i,b,t}\right)}\right\}\exp\left\{-\frac{\left(\mu_b-\mu_j-\frac{\delta_{\rm min}}{2}\right)^2 t}{2\sigma_b^2\left(1+\sum_{i\ne b}\underline{\rho}_{i,b,t}\right)}\right\}\rule{0em}{10mm}\right\}.
\end{align*}
\hfill$\blacksquare$

\section{Proof of Theorem \ref{thm3}}

According to \cite{Bubeck2012}, Page 9, 
\begin{align}\label{prop4-eq1}
R_t=\sum_{i\ne b}\left(\mu_b-\mu_i\right)\mathbb{E}\left[N_{i,t}\right].
\end{align}
Combining (\ref{thm1-expected-number-nb-up}) and (\ref{prop4-eq1}), 
\begin{align*}
R_t<\frac{\sum_{i\ne b}\left(\mu_b-\mu_i\right)\overline{\rho}_{i,b,t}}{1+\sum_{i\ne b}\underline{\rho}_{i,b,t}}t+k\sum_{i\ne b}\left(\mu_b-\mu_i\right)q\left(\frac{3}{4}t\right)t.
\end{align*}
According to Lebesgue's dominated convergence theorem, for $\forall i$,
\begin{align*}
\underset{t\to\infty}{\lim}\mathbb{E}\left[\frac{N_{i,t}}{t}\right]=\mathbb{E}\left[\underset{t\to\infty}{\lim}\frac{N_{i,t}}{t}\right].
\end{align*}
Based on Corollary \ref{cor1}, it holds that $\mathbb{P}\left(\lim_{t\to\infty}\alpha_{i,t}=\frac{\frac{\sigma_i}{\mu_b-\mu_i}}{\frac{\sigma_b}{\mu_b-\max_{j\ne b}\mu_j}+\sum_{i\ne b}\frac{\sigma_i}{\mu_b-\mu_i}}\right)=1$ for $\forall i\ne b$. Thus, 
\begin{align*}
&\lim_{t\to\infty}\frac{R_t}{t}\\
=&\sum_{i\ne b}\left(\mu_b-\mu_i\right)\mathbb{E}\left[\lim_{t\to\infty}\alpha_{i,t}\left|\lim_{t\to\infty}\alpha_{i,t}=\frac{\frac{\sigma_i}{\mu_b-\mu_i}}{\frac{\sigma_b}{\mu_b-\max_{j\ne b}\mu_j}+\sum_{i\ne b}\frac{\sigma_i}{\mu_b-\mu_i}}\right.\right]\\
&~~~~~\cdot\mathbb{P}\left(\lim_{t\to\infty}\alpha_{i,t}=\frac{\frac{\sigma_i}{\mu_b-\mu_i}}{\frac{\sigma_b}{\mu_b-\max_{j\ne b}\mu_j}+\sum_{i\ne b}\frac{\sigma_i}{\mu_b-\mu_i}}\right)\\
&+\sum_{i\ne b}\left(\mu_b-\mu_i\right)\mathbb{E}\left[\lim_{t\to\infty}\alpha_{i,t}\left|\lim_{t\to\infty}\alpha_{i,t}\ne\frac{\frac{\sigma_i}{\mu_b-\mu_i}}{\frac{\sigma_b}{\mu_b-\max_{j\ne b}\mu_j}+\sum_{i\ne b}\frac{\sigma_i}{\mu_b-\mu_i}}\right.\right]\\
&~~~~~~~~\cdot\mathbb{P}\left(\lim_{t\to\infty}\alpha_{i,t}\ne\frac{\frac{\sigma_i}{\mu_b-\mu_i}}{\frac{\sigma_b}{\mu_b-\max_{j\ne b}\mu_j}+\sum_{i\ne b}\frac{\sigma_i}{\mu_b-\mu_i}}\right)\\
=&\sum_{i\ne b}\frac{\left(\mu_b-\mu_i\right)\frac{\sigma_i}{\mu_b-\mu_i}}{\frac{\sigma_b}{\mu_b-\max_{j\ne b}\mu_j}+\sum_{i\ne b}\frac{\sigma_i}{\mu_b-\mu_i}}
=\frac{\sum_{i\ne b}\sigma_i}{\frac{\sigma_b}{\mu_b-\max_{j\ne b}\mu_j}+\sum_{i\ne b}\frac{\sigma_i}{\mu_b-\mu_i}}.
\end{align*}
\hfill$\blacksquare$

\section{Proof of Theorem \ref{thm4}}

Similar to the discussion in Section \ref{thm2-proof}, there exists a measurable set $\Omega_3\subseteq\Omega$ such that $\mathbb{P}\left(\omega\in\Omega_3\right)=1$, $\hat{\mu}_{i,t}\to\mu_i$ as $t\to\infty$, and $\hat{\mu}_{i_1,t_1}\ne\hat{\mu}_{i_2,t_2}$ for all $i_1\ne i_2$ and $t_1,t_2\geq 1$. Following similar discussion as used in (\ref{PE_new}), for $\forall t$, 
\begin{align*}
e_t=\mathbb{P}\left(\left.J_t\ne b\right|\omega\in\Omega_3\right).
\end{align*}
Under the KG algorithm, if the event $\mathcal{E}_n$ introduced in Section \ref{thm2-proof} occurs in round $n$, then the correct selection occurs in round $n$ regardless of the exact values of $N_{i,n}$'s. Following similar discussion in Section \ref{thm2-proof},
\begin{align}\label{en-upper-1}
e_n\leq\mathbb{P}\left(\left.\mathcal{E}_n^{c}\right|\omega\in\Omega_3\right)
\leq\mathbb{P}\left(\left.\hat{\mu}_{b,n}\leq\mu_b-\frac{\delta_{\rm min}}{2}\right|\omega\in\Omega_3\right)+\sum_{i\ne b}\mathbb{P}\left(\left.\hat{\mu}_{i,n}>\mu_b-\frac{\delta_{\rm min}}{2}\right|\omega\in\Omega_3\right).
\end{align}
Similar to the discussion in Section \ref{thm2-proof}, we can derive that if given $N_{i,n}=n_i$, $i=1,\dots,k$,
\begin{align}
\mathbb{P}\left(\left.\hat{\mu}_{b,n}\leq\mu_b-\frac{\delta_{\rm min}}{2}\right|\omega\in\Omega_3\right)\leq&\frac{\sqrt{2}\sigma_b}{\sqrt{\pi n_b}\delta_{\rm min}}\exp\left\{-\frac{\delta_{\rm min}^2}{8\sigma_b^2}n_i\right\},\label{thm4-PE_upper-1}\\
\mathbb{P}\left(\left.\hat{\mu}_{i,n}>\mu_b-\frac{\delta_{\rm min}}{2}\right|\omega\in\Omega_3\right)\leq&\frac{\sigma_i}{\sqrt{2\pi n_i}\left(\mu_b-\mu_i-\frac{\delta_{\rm min}}{2}\right)}\exp\left\{-\frac{\left(\mu_b-\mu_i-\frac{\delta_{\rm min}}{2}\right)^2}{2\sigma_i^2}n_i\right\}.\label{thm4-PE_upper-2}
\end{align}
Notice that $N_{i,n}\geq n_0$ for $\forall i$, $\forall n>kn_0$. By combining it with (\ref{en-upper-1}), (\ref{thm4-PE_upper-1}), (\ref{thm4-PE_upper-2}), and $n_0=\left\lfloor\alpha_0 n\right\rfloor$,
\begin{align*}
e_n\leq&\frac{\sqrt{2}\sigma_b}{\sqrt{\pi\left\lfloor\alpha_0 n\right\rfloor}\delta_{\rm min}}\exp\left\{-\frac{\delta_{\rm min}^2}{8\sigma_b^2}\left\lfloor\alpha_0 n\right\rfloor\right\}\\
&+\sum_{i\ne b}\frac{\sigma_i}{\sqrt{2\pi \left\lfloor\alpha_0 n\right\rfloor}\left(\mu_b-\mu_i-\frac{\delta_{\rm min}}{2}\right)}\exp\left\{-\frac{\left(\mu_b-\mu_i-\frac{\delta_{\rm min}}{2}\right)^2}{2\sigma_i^2}\left\lfloor\alpha_0 n\right\rfloor\right\}.
\end{align*}
In addition, if the event $\tilde{\mathcal{E}}_n$ introduced in Section \ref{prop3-proof} occurs in round $n$, then the false selection occurs in round $n$ regardless of the exact values of $N_{i,n}$'s. Following similar discussion in Section \ref{prop3-proof}, $\exists j\ne b$,
\begin{align}\label{en-lower-1}
e_n\geq\mathbb{P}\left(\left.\tilde{\mathcal{E}}_{j,n}\right|\omega\in\Omega_3\right)
\geq\mathbb{P}\left(\left.\hat{\mu}_{j,n}>\mu_j+\frac{\delta_{\rm min}}{2}\right|\omega\in\Omega_3\right)\mathbb{P}\left(\left.\hat{\mu}_{b,n}\leq\mu_j+\frac{\delta_{\rm min}}{2}\right|\omega\in\Omega_3\right).
\end{align}
Similar to the discussion in Section \ref{thm2-proof}, we can derive that if given $N_{i,n}=n_i$, $i=1,\dots,k$,
\begin{align}
\mathbb{P}\left(\left.\hat{\mu}_{j,t}>\mu_j+\frac{\delta_{\rm min}}{2}\right|\omega\in\Omega_3\right)\geq&\frac{\delta_{\rm min}\sqrt{n_j}}{2\sqrt{2\pi}\sigma_j\left(1+\frac{\delta_{\rm min}^2}{4\sigma_j^2}n_j\right)}\exp\left\{-\frac{\delta_{\rm min}^2}{8\sigma_j^2}n_j\right\},\label{thm4-PE_lower-1}\\
\mathbb{P}\left(\left.\hat{\mu}_{b,t}\leq\mu_j+\frac{\delta_{\rm min}}{2}\right|\omega\in\Omega_3\right)\geq&\frac{\left(\mu_b-\mu_j-\frac{\delta_{\rm min}}{2}\right)\sqrt{n_b}}{\sqrt{2\pi}\sigma_b\left(1+\frac{\left(\mu_b-\mu_j-\frac{\delta_{\rm min}}{2}\right)^2}{\sigma_b^2}n_b\right)}\exp\left\{-\frac{\left(\mu_b-\mu_j-\frac{\delta_{\rm min}}{2}\right)^2}{2\sigma_b^2}n_b\right\}.\label{thm4-PE_lower-2}
\end{align}
By combining $\left\lfloor\alpha_0 n\right\rfloor\leq N_{i,n}\leq n$ with (\ref{en-lower-1}), (\ref{thm4-PE_lower-1}) and (\ref{thm4-PE_lower-2}), 
\begin{align*}
e_n\geq\min_{j\ne b}&\left\{\frac{\delta_{\rm min}}{2\sqrt{2\pi}\sigma_j\left(1+\frac{\delta_{\rm min}^2}{4\sigma_j^2}n\right)}
\frac{\left(\mu_b-\mu_j-\frac{\delta_{\rm min}}{2}\right)\left\lfloor\alpha_0 n\right\rfloor}{\sqrt{2\pi}\sigma_b\left(1+\frac{\left(\mu_b-\mu_j-\frac{\delta_{\rm min}}{2}\right)^2}{\sigma_b^2}n\right)}\right.\\
&~~~~\left.\cdot
\exp\left\{-\frac{\delta_{\rm min}^2}{8\sigma_j^2}n\right\}
\exp\left\{-\frac{\left(\mu_b-\mu_j-\frac{\delta_{\rm min}}{2}\right)^2}{2\sigma_b^2}n\right\}\rule{0em}{12mm}\right\}.
\end{align*}
By following similar discussion in Sections \ref{thm2-proof} and \ref{prop3-proof}, we can derive that
\begin{align*}
r_n\leq&\frac{\sqrt{2}\sigma_b\delta_{\rm max}}{\sqrt{\pi\left\lfloor\alpha_0 n\right\rfloor}\delta_{\rm min}}\exp\left\{-\frac{\delta_{\rm min}^2}{8\sigma_b^2}\left\lfloor\alpha_0 n\right\rfloor\right\}\\
&+\sum_{i\ne b}\frac{\sigma_i\delta_{\rm max}}{\sqrt{2\pi \left\lfloor\alpha_0 n\right\rfloor}\left(\mu_b-\mu_i-\frac{\delta_{\rm min}}{2}\right)}\exp\left\{-\frac{\left(\mu_b-\mu_i-\frac{\delta_{\rm min}}{2}\right)^2}{2\sigma_i^2}\left\lfloor\alpha_0 n\right\rfloor\right\},
\end{align*}
\begin{align*}
r_n\geq\min_{j\ne b}&\left\{\frac{\delta_{\rm min}^2}{2\sqrt{2\pi}\sigma_j\left(1+\frac{\delta_{\rm min}^2}{4\sigma_j^2}n\right)}
\frac{\left(\mu_b-\mu_j-\frac{\delta_{\rm min}}{2}\right)\left\lfloor\alpha_0 n\right\rfloor}{\sqrt{2\pi}\sigma_b\left(1+\frac{\left(\mu_b-\mu_j-\frac{\delta_{\rm min}}{2}\right)^2}{\sigma_b^2}n\right)}\right.\\
&~~~~\left.\cdot\exp\left\{-\frac{\delta_{\rm min}^2}{8\sigma_j^2}n\right\}
\exp\left\{-\frac{\left(\mu_b-\mu_j-\frac{\delta_{\rm min}}{2}\right)^2}{2\sigma_b^2}n\right\}\rule{0em}{12mm}\right\}.
\end{align*}
\hfill$\blacksquare$

\end{document}